\documentclass[Afour,sageh,times]{sagej}

\usepackage{moreverb}
\usepackage{cite}
\usepackage{array}
\usepackage{amsmath,amssymb,amsfonts}
\usepackage{graphicx}
\usepackage{caption} %for subfigures
\usepackage{textcomp}
\usepackage{xcolor}
\usepackage{float}
\usepackage{tabu}
\usepackage{color}
\usepackage{footnote}
\usepackage{mathptmx}
\usepackage{upgreek}
\usepackage{gensymb}
\usepackage{mathrsfs}
\usepackage{siunitx}
\usepackage{algorithm}
\usepackage{algpseudocode}
\usepackage{tabularx} % for tables
\usepackage{multirow} % multirows in tables

\usepackage{subfig}
\graphicspath{{figs/}}
\usepackage[labelfont=bf]{caption}
\usepackage[utf8]{inputenc}
\usepackage{mathtools}
\usepackage{booktabs}

\usepackage[colorlinks,bookmarksopen,bookmarksnumbered,citecolor=gray,urlcolor=gray]{hyperref}

\usepackage{cleveref}

\DeclareMathOperator*{\argmin}{arg\,min}

\newcommand{\p}[1]{\medskip \noindent \textbf{{#1}.}}

\def\volumeyear{2026}
\def\journalname{The International Journal of Robotics Research}

\begin{document}

\runninghead{Fuentes et al.}

\title{Linking Exteroception and Proprioception through Improved Contact Modeling for Soft Growing Robots}

\author{Francesco Fuentes\affilnum{1},
        Serigne Diagne\affilnum{1},
        Zachary Kingston\affilnum{2}, and
        Laura H. Blumenschein\affilnum{1}}

\affiliation{\affilnum{1}School of Mechanical Engineering, Purdue University, West Lafayette, IN, USA\\
\affilnum{2}Department of Computer Science, Purdue University, West Lafayette, IN, USA}

\corrauth{Francesco Fuentes, School of Mechanical Engineering, Purdue University, West Lafayette, IN 47906, USA}
\email{ffuente@purdue.edu}

\begin{abstract}
Passive deformation due to compliance is a commonly used benefit of soft robots, providing opportunities to achieve robust actuation with few active degrees of freedom. Soft growing robots in particular have shown promise in navigation of unstructured environments due to their passive deformation. If their collisions and subsequent deformations can be better understood, soft robots could be used to understand the structure of the environment from direct tactile measurements. In this work, we propose the use of soft growing robots as mapping and exploration tools. We do this by first characterizing collision behavior during discrete turns, then leveraging this model
to develop a geometry-based simulator that models robot trajectories in 2D environments. Finally, we demonstrate the model and simulator validity by mapping unknown environments using Monte Carlo sampling to estimate the optimal next deployment given current knowledge. Over both uniform and non-uniform environments, this selection method rapidly approaches ideal actions, showing the potential for soft growing robots in unstructured environment exploration and mapping.
\end{abstract}


\maketitle

\section{Introduction}
Compliance within soft robots enables adaptability and robustness during physical interactions, particularly through passive deformation during actuation.
Examples include soft grippers that conform to object shapes during grasping and locomoting robot legs that deform over irregularities and bumps in the terrain.
These deformation events are usually treated as passive reactions to environmental uncertainty rather than sensed phenomena; however, these interactions contain rich information about the environment, the interaction itself, and the robot state that could be extracted through proper sensing and interpretation.

Realizing this potential requires models that translate internally sensed deformation into external environmental features and vice versa.
Recent work in tactile sensing demonstrates this concept, using passive deformation at the level of the full finger \citep{jamil2021proprioceptive,Thuruthel2019} and at the level of the contact surface \citep{athar2025vibtac,zhang2022hardware} to identify grasped objects and their positions.
While some applications employ first-principle models (e.g., cable orientation \citep{She2021}), many rely on machine learning to map deformation to object properties \citep{akinola2025tacsl,zhang2025design}.
These learning-based approaches limit transferability to novel applications where tactile sensing could provide significant benefits, such as exploration and mapping.

\begin{figure}[t!]
\centerline{\includegraphics[width=0.9\columnwidth]{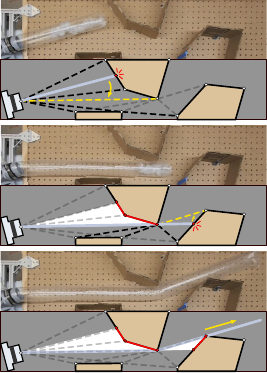}}
    \caption{Leveraging knowledge of passive kinematics for soft growing structures under contact, we can predict robot deformations and extract information about occupied and unoccupied regions of an environment, allowing uses in exploring and mapping unknown environments.}
    \label{fig:intro}
\end{figure}

For mapping applications, we require soft robot structures that undergo reliable sensing ``events'' during environmental contact to provide predictable transformations between internal deformation and external interaction.
A recent class of soft robots, vine robots (i.e., soft growing robots) meet these requirements through their unique combination of locomotion and manipulation behaviors \citep{blumenschein2020design}, enabling navigation through cluttered environments via growth and passive deformation \citep{Hawkes2017}.
Critically, this passive navigation follows relatively simple kinematic models based on contact angle and robot state \citep{Greer2018}, which enable robust path planning \citep{greer2020robust}.
We hypothesize that these same behaviors can be leveraged for environmental mapping using only passive steering (Figure~\ref{fig:intro}).

To that end, in this paper we present the first demonstration that environment mapping in an unknown environment can be done with a passive soft growing robot with limited sensing, though we note that sensor integration remains a future task. Overall, we make the following contributions:

\p{Expansion of Passive Kinematics Models} We expand existing models of passive kinematics under obstacle contact to include static pre-formed turns. Working from first principles of inflated beam mechanics, we identify four distinct behaviors resulting from passive compliance when a turn is added to a soft growing robot. These behaviors can be robustly predicted and exploited to allow the robot to access more areas of an unknown environment.

\p{Simulation of Repeated Obstacle Interactions} We develop a simulator which implements the passive steering kinematic equations and predicts the path of growing robots through multi-object interactions in cluttered environments. This simulation leverages visibility graphs of the environment to precisely identify the final shape of the soft growing robot.

\p{Algorithm to Map Unknown Environments} Using a sampling-based Monte Carlo approach, we demonstrate the use of soft growing robots with passive responses to successfully map an environment. Starting with no knowledge of an environment, we sample potential environments from occupancy probability grids to make informed choices of the best robot to send next into the true environment.

\p{Demonstration of Mapping Using a Physical Robot Deployment} We demonstrate the above mapping approach in a real environment with external cameras acting in place of internal sensors. We use only extracted tip contact angle and robot length as noisy data representative of current sensing methods. We show that passive robot kinematics can gather environment data using limited sensor information and only a few robots to accurately estimate the environment.

\subsection{Related Works}

\subsubsection*{Touch Sensing}
%haptic
Of the human senses, touch is the only one to require an active process and a physical interaction \citep{Prescott2011}. Through touch, we gain many insights into an object's physical properties, such as friction and compliance, which are important to consider when interacting with an object. Specifically, this sense is informed through understanding the deformations in the interface of the physical interaction (e.g., as demonstrated in \citet{Hoepflinger2010} to detect terrain). 
This directly leads to haptic sensing demanding a degree of compliance in order to actually be informed successfully by touch.

Detecting these deformations can occur through a variety of mediums.
Vision-based approaches have gained popularity due to their resolution, exemplified by GelSight sensors \citep{Yuan2017,She2021}, which use embedded cameras to capture elastomer deformation and then reconstruct the geometry.
These sensors can measure shear and normal forces, detect textures, and measure slip \citep{Yuan2015}. Alternative approaches have used infrared sensing \citep{Patel2018}, embedded sensors in fingers \citep{Shih2017}, and bio-inspired sensor arrays such as the star-nose-inspired tactile-olfactory system \citep{Liu2022}. A more extensive review of soft tactile sensing can be found in \citet{Roberts2021}. Still, tactile sensing in this way forces specific areas of contact in order to sense exterior features and struggles to expand to general touch along the robot.

\subsubsection*{Proprioceptive Sensing}
% proprioceptive
Proprioception is the ability to be aware of one's physical position and movement in space \citep{Proske2012}, an important quality for animals, or robots, to robustly navigate a space. The difference between the expected positions of one's body, and what proprioception senses, informs interactions with external objects. As an example, attempt to (lightly) poke your finger through a flat surface with your eyes closed. Whereas one's haptic sense informs \textit{what} the surface is, one's proprioception informs \textit{where} it is in relation to your body, purely from the `buckling' of the finger as it makes contact.

This type of internal sensing can be greatly beneficial to the control and navigation of a robot. For example, several works by McGarey et al. take advantage of the tension and orientation in a tether between wheeled robots to localize both the robots and any objects that may have been wrapped around by the tether during navigation \citep{McGarey2016,McGarey2017}.  \citet{Wang2020snake} used proprioception in a serpentine robot to improve its locomotion by pushing against parts of its environment.

For soft robots, proprioceptive sensing presents a unique challenge, as sensors can affect the already uncertain robot dynamics and must deform with the robot \citep{Thuruthel2019}. Some have tackled this approach with bio-inspiration, such as rat-inspired whiskers with strain gauges \citep{Pearson2005} and caterpillars that use friction-induced deformations to adapt their gait \citep{Umedachi2016}.
Recent work in stretchable sensors addresses this by deforming \textit{with} the robot while maintaining sensing capabilities \citep{White2017}.
Other examples include stretchable fiber optics \citep{Scharff2021}, 3D printed sensors \citep{Heiden2022} and magnetic sensor arrays to sense shape \citep{Baaij2023} and localize \citep{Watson2020}. Note, most current approaches use exteroceptive methods to inform proprioceptive models (e.g., using learned models from camera data \citep{Soter2018}). Our work expands this paradigm, using kinematic models to link proprioceptive sensing and exteroceptive sensing, allowing an understanding of the robot shape to yield an understanding of the environment and vice versa.

\subsubsection*{Vine Robots}

Vine robots, also known as soft growing robots, represent a recent class of soft continuum robots characterized by tip eversion, significant length changes, and passive environmental interaction guiding growth \citep{Hawkes2017}. Through this, these systems can navigate through cluttered environments by exploiting contact with obstacles rather than avoiding them, achieving better results than pure avoidance \citep{Greer2018, greer2020robust}. Additionally, the behavior of these systems when interacting with the environment have been characterized in \citet{Haggerty2019}. These works laid the basis for geometric prediction of robot behavior when interacting with obstacles, which we build upon. There have also been works that explore dynamic simulation of vine robots with obstacle interactions by using an approximation of the robot as a serial chain with impulse-based contact dynamics \citep{Jitosho2021, chen2025physics}.

The actuation of vine robots can be achieved in a number of ways, such as steering via pneumatic muscles~\citep{Coad2020}, helical actuation \citep{blumenschein2018}, and tendon routing \citep{blumenschein2022}. They can also be used as sensors, such as in \citet{frias2023} to localize via collision, and in \citet{gruebele2021distributed} as a distributed sensor network. Most similar to our work is that of \citet{wu2025soft}, which used proximity sensors and trial-and-error pre-bending to explore. In contrast, our work derives models of bending and deformation from first principles and applies them to intelligently select pre-bent robots for mapping. Prior work showed a feasible upper bound on this type of exploration assuming ideal environment knowledge \citep{Fuentes2023}.

Field deployments have validated vine robot capabilities in real-world scenarios, including archaeological exploration \citep{Coad2020} and urban search and rescue operations \citep{mcfarland2024field}. These applications demonstrate the potential for vine robots to serve as both mapping and sensing platforms in environments where traditional robots are unable to go.

\section{Kinematic Modeling of Passive Deformation as Sensing}\label{sect:equations}

Vine robots have exhibited beneficial behaviors when interacting with obstacles, specifically when it comes to navigating constrained environments \citep{Greer2018}. To begin addressing how vine robots can sense external environment features, we must define how the kinematics, i.e. the passive deformations of the robot as it grows, develop as a result of the environment.
In this section we will first review the existing models of vine robots growing into environments and passively deforming under the contact. Then we will expand those models to include kinematics with preplanned turns and describe how these models can be inverted to sense features of the environment.

\subsection{Kinematics for Straight Vine Robots}

\begin{figure}[t!]
\centerline{\includegraphics[width=1\columnwidth]{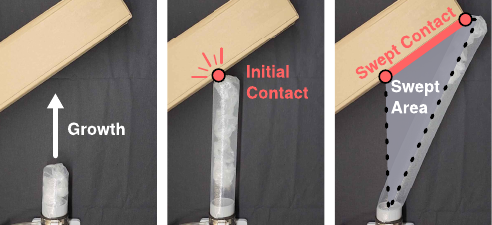}}
\caption{Vine robot movement under contact with the environment can be predicted based on the initial contact geometry and friction, and will pivot about a previous point while the tip follows the wall. The motion can generate two pieces of information, the wall location through swept contact, and free space, through swept area.}
\label{fig:VR_contact}
\end{figure}

When the growth direction of an unsteered vine robot takes it into contact with an obstacle, we can observe a robust and relatively simple-to-describe behavior as it continues to grow as seen in Figure~\ref{fig:VR_contact}: (1) the tip of the vine robot will move tangent to the obstacle surface and (2) the inflated tube that forms the vine robot's body will pivot (i.e. bend) at the most distal point of contact with the environment \citep{Greer2018,greer2020robust}. The robot will continue to pivot about this localized point as it continues to increase in length, until the tip reaches the corner of the obstacle. At this point, the robot will extend in the direction of its current orientation, until another obstacle is encountered, with the previous obstacle contact becoming a potential point to pivot about. This behavior can be seen in the real shots through Figure~\ref{fig:intro}. As described by \citep{greer2020robust}, the motion of the vine robot can therefore be defined by the tangents of the walls and a series of pivot points where obstacles contact the robot along its body, each of which can support either a positive (i.e. counter-clockwise) reaction moment or a negative (i.e. clockwise) reaction moment. If contact is made to both sides of the robot, pivots in either direction can be achieved at that point.

While this initial work geometrically describes where the vine robot is most likely to go as it grows, to fully understand the kinematics we need to understand the force balance underlying the behavior. Looking at the concentrated bending behavior at the pivot, we can note this arises due to the force from the contact at the tip causing a reduction in the material tension opposite the contact, as described by \citet{comer63} and \citet{leonard60}. The maximum internal restoring moment, $M_{int}$, from the inflated beam will be:
\begin{equation}\label{Eq:internalMoment}
    M_{int} = \pi P R^3
\end{equation}
where \(P\) is the internal pressure and \(R\) is the tube radius. However, this is not the only way the inflated tube can bend. If the tip is restricted from moving along the wall, e.g. due to friction, the inflated beam will experience beam buckling after a critical axial force, \(F_{buckling}\), is surpassed \citep{fichter1966}:
\vspace{-0.1cm}
\begin{equation}\label{axial_buckling}
F_{buckling} = \dfrac{E \pi^3 R^4 t P + E G \pi^3 R^3 t^2}{E \pi^2 R^2 t + R L^2 P + G t L^2}
\end{equation}
where $E$ and $G$ are the elastic modulus and shear modulus of the wall material, respectively, $t$ is the thickness of the tube material, and $L$ is the unsupported length of the robot body.

\begin{figure}[t!]
\centerline{\includegraphics[width=1\columnwidth]{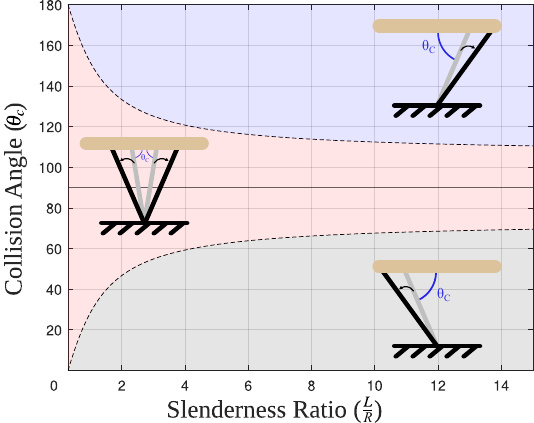}}
\caption{Predicted regions of contact behavior with a straight vine robot growing into a flat wall. Dashed lines plot the critical contact angle between pivoting and axial buckling behaviors as a function of slenderness ratio using \(\mu = 0.3\), \Cref{Eq:straightVR}.}
\label{fig:Og_coll_thresh}
\end{figure}

When vine robots contact obstacles they will axially buckle if the force required to overcome both the friction at the point of contact and the restoring moment of the tube due to pressure is larger than the axial force required to buckle. Given that both tangential and axial force components result from pressure, a force balance can be derived to predict the critical angle at which a vine robot switches from buckling to bending when growing into a wall, as first shown by \citep{Haggerty2019}. The resulting critical threshold for the contact angle is:
\begin{equation}\label{Eq:straightVR}
    \theta_c^* = \tan^{-1}\left({\frac{\dfrac{L}{R}-\mu}{\mu \dfrac{L}{R}+1}}\right)
\end{equation}
where $\theta_c^*$ is the critical angle as measured relative to the surface tangent, $\mu$ is the static friction coefficient, and $L$ is the length from the wall contact to the previous pivot point. %, and $R$ is the robot radius. 
The equation can be non-dimensionalized by the slenderness ratio, $L/R$. The boundary between  sliding wall contact and the axial buckling region can largely be determined by the geometry alone as shown by Figure~\ref{fig:Og_coll_thresh}. If we keep a consistent direction of measurement for the contact angle of the robot with the wall, we can reflect this critical angle about the $90^{\circ}$ line, where contact is perpendicular to the wall, thereby getting both positive and negative rotations.

While the critical angle can be rather large for low slenderness ratios, it quickly moves towards $90^{\circ}$ as the slenderness ratio increases. For most slenderness ratios we can expect axial buckling within $10$ to $15^{\circ}$ of perpendicular. Outside this critical region, we can easily predict the behavior under collision merely from the contact angle. Within the critical region, there is more uncertainty, but generally the behavior will be biased towards whichever of the pivoting directions it is closer to, as the axial buckling will be more likely to trend in this direction, though with the buckled point at a different location. 

\begin{figure*}[h!]
\centerline{\includegraphics[width=1\linewidth]{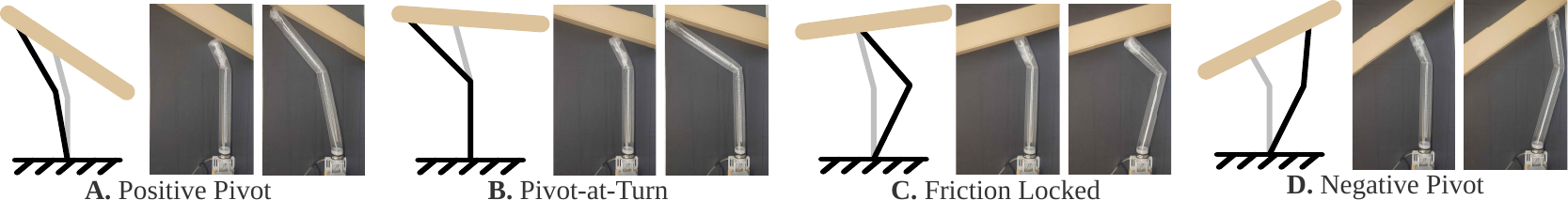}}
\caption{The four post-collision morphologies that occur when a vine robot with a turn grows into a wall, each with a physical demonstration of its movement. A) Positive Pivot, B) Pivot-at-Turn, C) Friction-Locked, and D) Negative Pivot.}
\label{fig:morphologies}
\end{figure*}

\subsection{Kinematics for Non-Straight Vine Robots}

The kinematics for unactuated vine robots provide useful insights for how environment contacts can shape the robot, but the use of them in sensing environments can be limited by inaccessible regions, since only the launch location and orientation can be set \citep{Fuentes2023}. To expand the control space, we add steering to the vine robots path using discrete, permanent turns. While vine robots have been actuated in a number of ways that result in different steered shapes, from continuous steering with pneumatic muscles or tendons~\citep{Coad2020}, to distributed discrete steering with latches \citep{Hawkes2017}, to serial actuation to create complex shapes \citep{blumenschein2022, wang2024}, discrete turns  produce a consistent angle and therefore can respond consistently to the geometry of the environment.

To create a discrete-type turn, a concentrated buckling is induced at one section of the vine robot. This can be done in a variety of ways, but the simplest is done during the construction of the robot; by folding the material and taping it down, one side of the robot body can be made shorter than the other \citep{agharese2023configuration}. As the vine robot grows, the fold forces a localized turn at that point along the length of the robot. Depending on the length of the fold, one can obtain different angles of turning when the robot is deployed.

When considering expanding the geometry based kinematics model of the vine robot, these discrete-type turns offer an ideal steering addition, as they are localized and, at a high level, act similarly to environment based pivot points, as shown previously by \citet{greer2020robust}, which treated the turns as pivot points with the same kinematics as environment contacts. However, in practice, it was found that this is not always the case, and that the critical angle predictions for straight robots from Equation~(\ref{Eq:straightVR}) do not accurately predict the behaviors with turns.

We can derive the factors which influence the kinematic behaviors by building on our understanding of the straight inflated tube kinematics. The previous behaviors result from: 1) the inflated tube restoring moment being independent of bend angle \citep{chen2025physics}, so turns will be localized to a single point, 2) the relative magnitude and direction of bending loads depending only on contact angle, and 3) contact friction dominating the transition between clockwise bending and counterclockwise bending. For a vine robot with a single bend before contact, these features result in four potential behaviors upon contact, as shown in Figure~\ref{fig:morphologies}: Positive Pivot, Pivot-at-Turn, Friction-Locked, and Negative Pivot. If the contact force overcomes friction, the robot can pivot at the contact with the environment or at the turn location (only in the direction of the turn). Which of these occurs depends on which direction the moment acts and which point has a higher bending moment from the force at the tip. If instead friction dominates, the end of the robot remains in place but the robot continues to grow and will bend at both the turn and the contact point. This behavior will resolve once the contact angle has shifted sufficiently to reduce the friction and allow the tip to move.

\begin{figure}[h!]
\centerline{\includegraphics[width=0.5\linewidth]{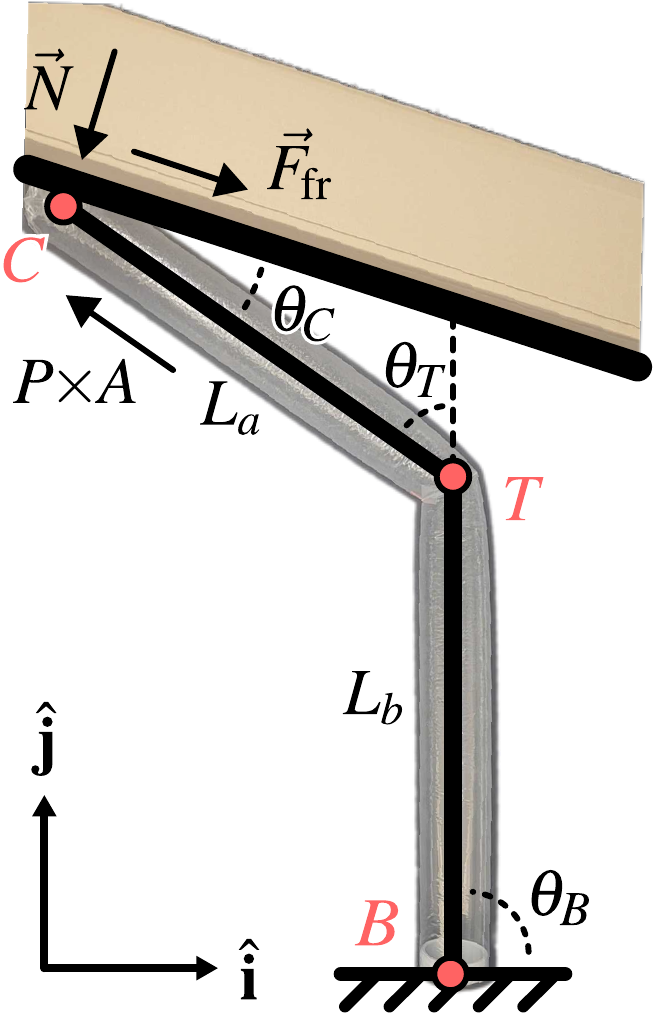}}
\caption{Notation used within the passive kinematic model for vine robots with a turn, including forces, robot and contact geometry, and labels for important points. Cartesian frame and robot orientation used in derivation is shown as well.}
\label{fig:notation}
\end{figure}

To calculate which morphology will occur, we first construct the moment balances for the environment contact point, $\vec{B}$, and the turn point, $\vec{T}$, based on the internal restorative moment from Equation~(\ref{Eq:internalMoment}), $M_{int}$, and both the normal, $\vec{N}$, and frictional, $\vec{F}_{fr}$, forces at the tip contact point, $\vec{C}$. Due to the robot's low mass and relatively slow tip growth, we can assume quasi-static conditions. Moreover, it has been consistently shown that the restoring moment is related only to geometric properties for thin walled inflated beams \citep{Haggerty2019,wang2025anisotropic,chen2025physics}, enabling the analysis to be a function of the robot geometry and environment features, the variables of which are shown in Figure~\ref{fig:notation}: $\theta_C$, the contact angle between the robot heading and the obstacle tangent, $\theta_T$, the turn angle between the heading before and after the turn, $\mu$, the friction coefficient of the contact, $L_b$, the length of the robot before the turn, $L_a$, the length of the robot after the turn, and $R$, the radius of the robot. Additionally, we define $\mathcal{L}$ as the ratio of the length before and after the turn, $\mathcal{L}=L_a/L_b$.

Without loss of generality, we assume that the turn angle, $\theta_T$, is positive and the first segment of the robot points in the positive $\hat{\textbf{j}}$ direction, as shown in Figure~\ref{fig:notation}. Starting with the moment about point $\vec{B}$, the moment balance will be:
\begin{equation}
   \Sigma\vec{M}_{B}=\vec{M}_{B,N}+\vec{M}_{B,F_{fr}}+\vec{M}_{int}
\end{equation}
Assuming that the robot is at the critical moment balance, i.e. it is in a quasi-static equilibrium right before moving, the moment due to friction and the internal restorative moment will counteract the moment due to the normal force. That moment from $\vec{N}$ can be calculated as:
\begin{equation}
\begin{split}
    \vec{M}_{B,N}&=\vec{r}_{B/C} \times\vec{N}= \begin{bmatrix}-L_a\sin{\theta_T}\\ L_b+L_a\cos{\theta_T}\\0\end{bmatrix} \times \begin{bmatrix}N\cos({\theta_T+\theta_C})\\N\sin({\theta_T+\theta_C})\\0\end{bmatrix}\\ &=N(L_a\cos({\theta_C})+L_b\cos({\theta_T+\theta_C})) \hat{\textbf{k}}
    \end{split}
\end{equation}
If this moment is positive, the friction will act in a direction to cause a negative moment and vice versa. Similarly, the internal restorative moment will have the opposite sign of the moment due to the normal force. We can write both critical moment cases together as:
\begin{equation}\label{Eq:sum_mom_o}
\begin{split}
    \Sigma M_B =&~ \vec{r}_{B/C} \times\vec{N}+\vec{r}_{B/C} \times\vec{F_{fr}} \pm P\pi R^3 \\
    =&N[L_a\cos({\theta_C})+L_b\cos({\theta_T+\theta_C})]\\ &\pm \mu N[L_a\sin({\theta_C})+L_b\sin({\theta_T+\theta_C})] \pm P\pi R^3=0
\end{split}
\end{equation}
The critical moment balance at point $T$ can be calculated similarly, but in this case we only consider where the normal force causes a positive moment, as a negative moment will not cause bending about the turn point:
\begin{equation} \label{EQ:sum_momm_T}
\begin{split}
    \Sigma M_T =& \vec{r}_{T/C} \times\vec{N}+\vec{r}_{T/C} \times\vec{F_{fr}} - P\pi R^3\\ =& NL_a\cos({\theta_C})-\mu N L_a\sin({\theta_C})-P\pi R^3=0
\end{split}
\end{equation}
Lastly, we can write the force balance at the tip:
\begin{equation}
\begin{split}
    \Sigma \vec{F}_C =& \vec{N}+\vec{F}_{fr}+\vec{PA}\\
    =& \begin{bmatrix}
        -N\cos({\theta_C+\theta_T})\pm\mu N (-\sin({\theta_C+\theta_T}))-PA\sin({\theta_T})\\
         - N\sin({\theta_C+\theta_T})\pm\mu N \cos({\theta_C+\theta_T})+PA\cos({\theta_T})
    \end{bmatrix}
    \end{split}
\end{equation}
Depending on the direction of friction, two solutions for the magnitude of $\vec{N}$ occur. When friction causes a positive moment at $\vec{O}$, the magnitude of the normal force will be:
\begin{equation} \label{Eq:pos_normal_force}
    N = \frac{P\pi R^2}{\sin\theta_C - \mu\cos\theta_C}
\end{equation}
and conversely, when friction causes a negative moment, the magnitude of the normal force will be:
\begin{equation} \label{Eq:neg_normal_force}
    N = \frac{P\pi R^2}{\sin\theta_C + \mu\cos\theta_C}.
\end{equation}

With these critical moment and force equations established, we can determine the barriers between the kinematic morphologies. Starting with Positive Pivot and Pivot-at-Turn, this boundary is defined by the moments at $\vec{B}$ and $\vec{T}$ being equal to each other. Combining Equations~(\ref{Eq:sum_mom_o}) and (\ref{EQ:sum_momm_T}) yields:
\begin{equation}
    \cos(\theta_T + \theta_C) - \mu\sin(\theta_T + \theta_C) =0
\end{equation}
Solving for \(\theta_C\) results in the following expression for the critical contact angle separating Positive Pivot (PP) from Pivot-at-Turn (PaT):
\begin{equation}\label{Eq:lean_static}
    \theta^*_{C,~\text{PP}|\text{PaT}} = \tan^{-1}\left(\frac{\cos\theta_T - \mu \sin\theta_T}{\sin\theta_T + \mu \cos\theta_T}\right)
\end{equation}
or alternatively:
\begin{equation}
   \label{Eq:lean_static_alt}
   \theta^*_{C,~\text{PP}|\text{PaT}} = \tan^{-1}\left(\frac{1}{\mu}\right) - \theta_T
\end{equation}

To separate Pivot-at-Turn from Friction-Locked (FL), we substitute Equation~(\ref{Eq:neg_normal_force}) into Equation~(\ref{Eq:sum_mom_o}), resulting in:
\begin{equation}\label{Eq:static_squish}
    \theta^*_{C,~\text{PaT}|\text{FL}} = \tan^{-1}\left(\frac{\mathcal{L}(\cos\theta_T - \mu \sin\theta_T) + 1 -\frac{\mu R}{L_a}}{\mathcal{L}(\sin\theta_T + \mu \cos\theta_T) + \mu + \frac{R}{L_a}}\right)
\end{equation}
Similarly, to separate Negative Pivot (NP) from Friction-Locked (FL), we substitute Equation~(\ref{Eq:pos_normal_force}) into Equation~(\ref{Eq:sum_mom_o}), resulting in:
\begin{equation}\label{Eq:squish_lean}
   \theta^*_{C,~\text{FL}|\text{NP}} = \tan^{-1}\left(\frac{\mathcal{L}(\cos\theta_T + \mu \sin\theta_T) + 1 -\frac{\mu R}{L_a}}{\mathcal{L}(\sin\theta_T - \mu \cos\theta_T) - \mu - \frac{R}{L_a}}\right)
\end{equation}

\begin{figure}[t!]
\centerline{\includegraphics[width=1\columnwidth]{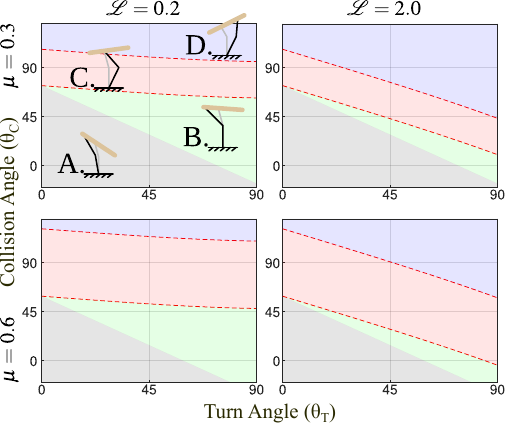}}
\caption{Slices of the 3D space divided by the critical angles for vine robots with turns given by~\Cref{Eq:lean_static_alt,Eq:static_squish,Eq:squish_lean}. Resulting behaviors in each region are marked by morphologies shown in Figure~\ref{fig:morphologies}. Rows use $\mu$ values of 0.3 (Top) and 0.6 (Bottom). Columns show length ratio, $\mathcal{L}$, values of 0.2 (Left) and 2 (Right).}
\label{fig:slices}
\end{figure}

If we assume a high slenderness ratio after the turn, i.e. $R<<L_a$, and a constant value for $\mu$, then each critical angle equation can be written as a function of just $\mathcal{L}$ and $\theta_T$. This results in a 3D surface (Figure~\ref{fig:morph_exp}), shown as slices of the volume for two different friction coefficients in Figure~\ref{fig:slices}.

As shown in Figure~\ref{fig:slices}, altering the length ratio affects the angle of the two boundaries with the Friction-Locked region and altering the static friction coefficient affects the thickness of the Friction-Locked region, with an increasing coefficient showing a thicker region. This relationship matches intuition, as more friction in wall contact will also increase the range of angles in which the vine robot's tip will not slide immediately. Increasing both these parameters, as shown in the bottom-right 2D slice of Figure~\ref{fig:slices}, both angles downward and expands the Friction-Locked region, making it much more likely to be encountered.

\subsubsection{Summary and Discussion}
Through a planar analysis of moments in a quasi-static interaction of a turning vine robot, three post-collision states and one collision state can be derived. These states are predominantly predicted by the moments generated from the internal pressure of the robot and the normal and friction forces at the tip, where contact occurs. Table~\ref{table:eq_review} summarizes the critical angle equations and the conditions at the boundary between collision states. Through these equations, one can predict the state of the robot as it continues growing and sliding against an obstacle of varying collision angle. As the collision angle increases, the collision state will transition through Positive Pivot, Pivot-at-Turn, Friction Locked, and Negative Pivot (Figure~\ref{fig:morphologies}).

Although we demonstrate the planar case, these equations could be extended to the third dimension in the cases where the obstacle is inclined away from the robot and it subsequently grows upwards. This would require consideration of object gradients as opposed to a single contact angle, though one could likely continue using the planar equations with a rotated reference frame; however, the torsional moments of the body and the robot's weight would have to be accounted for, so a full extension to 3D remains future work.

\begin{table*}[t!]
\caption{Summary of critical angle equations for non-straight vine robots in collision}
\label{table:eq_review}
\begin{center}
\resizebox{\textwidth}{!}{
\begin{tabular} { l|l|l }
\textbf{Positive Pivot \& Pivot-at-Turn} & \textbf{Pivot-at-Turn \& Friction Locked} & \textbf{Friction Locked \& Negative Pivot}\\
\hline
& &\\
\vspace{-0.2cm}
$\theta^*_{C,~\text{PP}|\text{PaT}} = \tan^{-1}\left(\frac{\cos\theta_T - \mu \sin\theta_T}{\sin\theta_T + \mu \cos\theta_T}\right)$ & $\theta^*_{C,~\text{PaT}|\text{FL}} = \tan^{-1}\left(\frac{\mathcal{L}(\cos\theta_T - \mu \sin\theta_T) + 1 -\frac{\mu R}{L_a}}{\mathcal{L}(\sin\theta_T + \mu \cos\theta_T) + \mu + \frac{R}{L_a}}\right)$ & $\theta^*_{C,~\text{FL}|\text{NP}} = \tan^{-1}\left(\frac{\mathcal{L}(\cos\theta_T + \mu \sin\theta_T) + 1 -\frac{\mu R}{L_a}}{\mathcal{L}(\sin\theta_T - \mu \cos\theta_T) - \mu - \frac{R}{L_a}}\right)$ \\
& &\\
\hline
Boundary occurs when the critical & Boundary occurs when the normal force to& Boundary occurs when the normal force to\\
moments at pivot and turn occur at&break sliding friction equals the normal force & break sliding friction equals the normal force\\
the same normal force& needed for the critical moment at the turn & needed for the critical moment at the pivot\\
 & (Friction causing a negative moment)& (Fiction causing a positive moment)\\
\end{tabular}
}
\end{center}
\end{table*}

\subsection{Validation of Turning Kinematics}
\begin{figure}[t!]
\centerline{\includegraphics[width=.9\columnwidth]{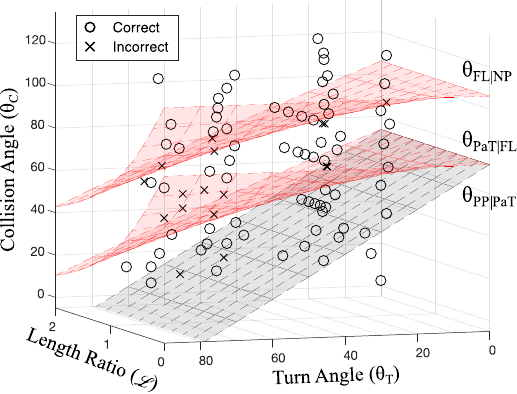}}
\caption{Experimental results for predicting the morphologies plotted onto the 3-dimensional region sliced by the critical contact angle surfaces (\Cref{Eq:lean_static_alt,Eq:static_squish,Eq:squish_lean})}
\label{fig:morph_exp}
\end{figure}

To test our expansion of the kinematic model for predicting vine robot motion, the morphology predictions in \Cref{Eq:lean_static_alt,Eq:squish_lean,Eq:static_squish} were tested by manufacturing various vine robots with different turning angles, $\theta_T$, and length ratios, $\mathcal{L}$. In each test, the vine robot was retracted and then pressurized to grow into an adjustable wall, similar to the examples shown in Figure~\ref{fig:morphologies}. These collisions were repeated for varying wall angles while the robot itself was launched from the same position and angle, thus only varying the collision angle, $\theta_C$ and \(\mathcal{L}\) due to slight changes in \(L_a\). Figure~\ref{fig:morph_exp} shows each test that was performed and whether the resulting behavior matched our prediction. The surfaces are plotted using a static friction coefficient value of 0.3, which was determined experimentally. The equations are fairly accurate in predicting when the morphologies occur, having an accuracy of 81.72\%. If we calculate the change in collision angle needed to put the behavior in the correct region, we get an average error of only 5.76 degrees. We can see that many of these errors occur at higher turn angle values, indicating this may be a region where the high slenderness ratio assumption is not as accurate.

\subsection{Kinematics as Sensing Expansion}\label{sect:kin_sens}

With this knowledge of how the kinematics of vine robot evolve as a function of the contact geometry, we can now use the derived principles to explain how the robot shape and the environment can be recovered from only simple information about the robot and the contact. While multiple different measurement sets could work, we only need to sense the contact angle and the change in robot length to recover both environment and robot shapes relative to the initial launch angle and position. While specific sensor design is outside the scope of this paper, similar tip-based sensing has been achieved previously as shown by~\citep{frias2023}. Instead, here we present the equations that would relate this sensed information to recover the robot and environment features.

For a straight vine robot segment, we know it will grow in the direction it is pointing until it hits something. %can determine where in space the wall is, and what its orientation is. 
Since we know the initial orientation of the vine robot, at first impact, we can determine where the collision is in space, $\vec{C_o}$, via:
\begin{equation}
    \vec{C_o} = \vec{B} + L\begin{bmatrix}
  \cos{\theta_B} \\
  \sin{\theta_B}
  \end{bmatrix}
\end{equation}
where \(\vec{B}\) is the previous pivot point that caused the body to bend, \(L\) is the vine robot length from the previous pivot point to the collision point, and \(\theta_B\) is the angle of the vine robot body from the pivot point to the contact point. Before any collisions occur, the first \(\theta_B\) is the initial launch angle. This term only gets updated once \(\vec{B}\) is updated.

As the vine robot tip slides against the wall, we can plot the point of contact, $\vec{C}_i$, using measurements of contact angle to update orientation. The equation is:
\begin{equation}\label{eq:str_wall_guess}
    \vec{C_i} = \vec{B} + L_i\begin{bmatrix}
  \cos(\theta_B+\theta_{C,0}-\theta_{C,i}) \\ 
  \sin(\theta_B+\theta_{C,0}-\theta_{C,i})
\end{bmatrix}
\end{equation}
where \(\theta_{C,0}\) is the collision angle measured at $\vec{C_o}$, and \(\theta_{C,i}\) is the $i$th collision angle measurement.

\begin{figure}[t!]
\centerline{\includegraphics[width=1\columnwidth]{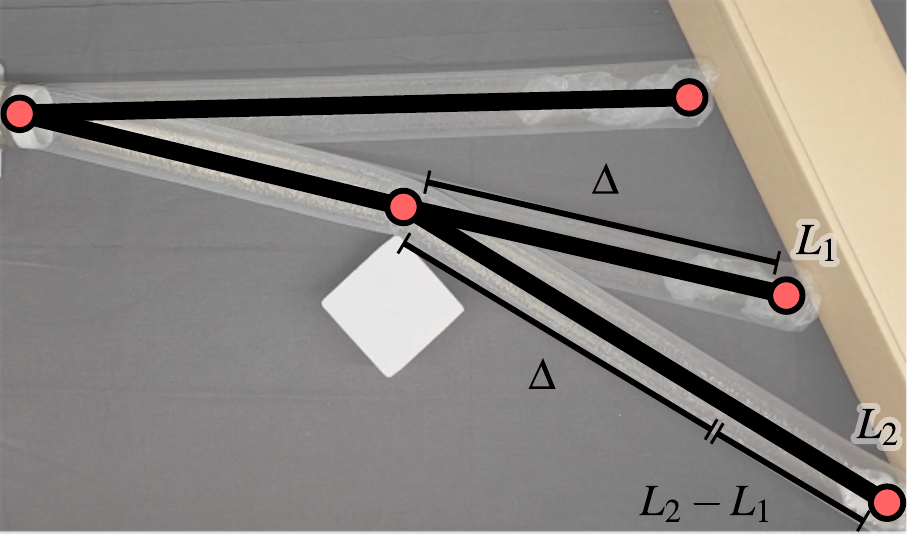}}
\caption{Variables to determine the location of a pivot change, derived purely from tip collision angle and robot length.}
\label{fig:pivot_coll}
\end{figure}

The previous equation will hold true as long as the pivot point remains the same. However, as the vine robot's tip slides along the face of an obstacle wall, its body may collide with another obstacle, forcing it to wrap around the obstacle (Figure~\ref{fig:pivot_coll}). This creates a new pivot point, $\vec{B}'$, and we can notice the change by the rate of change of the collision angle. With the assumption that the obstacle walls are flat, as the tip slides along the face of a wall and then encounters a new pivot, there will be a discontinuity in the derivative of the collision angle measurements relative to robot length. This discontinuity determines \textit{when} the new bend occurs. One can determine \textit{where} the new pivot point, \(\vec{B}'\), is located by the equation below:
\begin{equation}\label{eq:update_pivot}
    \vec{B}' = \vec{B} + (L-\Delta)\begin{bmatrix}
  \cos{\theta_B} \\ 
  \sin{\theta_B}
\end{bmatrix}
\end{equation}
where \(\Delta\) is distance from the new pivot point to the wall contact point when the robot first encountered the object, and is given by:
\begin{equation}\label{eq:len_from_pivot} 
    \Delta = \frac{(L_{i}-L_o)\cos{\theta_{C_{i}}}}{\sin{\theta_{C_o}}-\cos{\theta_{C_{i}}}}
\end{equation} 
where \(L_n\) is the vine robot length and \(\theta_{C_n}\) is the collision angle, and where subscript `o' is when the pivot is first encountered and `i' is some later point in time. Importantly, unless otherwise noted, all variable marked with an apostrophe to note they are updating that value replace the old value after calculation. Once the new pivot point is calculated, the estimation of the wall contact can be continued using the updated parameters.

If there is a turn point in between contacts, we can derive similar equations to calculate the shape information. Here we need the additional information of where along the vine robot body the turn is being induced, and the angle the turn creates. Regardless of the resulting morphology, we can calculate where the first point of contact, \(\vec{C_o}\), is with the following equation:

\begin{equation}\label{eq:contact_from_th_B}
    \vec{C_o} = \vec{B} + L_b\begin{bmatrix}
  \cos{\theta_B} \\ 
  \sin{\theta_B}
\end{bmatrix}+L_a\begin{bmatrix}
  \cos(\theta_B + \theta_T) \\ 
  \sin(\theta_B + \theta_T)
\end{bmatrix}.
\end{equation}

We can also calculate the wall's orientation immediately upon contact. To stay consistent, we assume the wall angle is always corrected to be within 0-180:
\begin{equation}\label{eq:wall_angle}
    \theta_W = \theta_C + \theta_T + \theta_B
\end{equation}
With the information of the geometry at contact, we next need to determine the morphology due to collision dynamics using~\Cref{Eq:lean_static_alt,Eq:static_squish,Eq:squish_lean}. Depending on the predicted morphology, there will be different methods to determine the robot and environment shape over the growth. For Pivot-at-Turn this is simply using the turn point as the starting pivot location in the previous equations, and then continuing to estimate the wall via~\Cref{eq:str_wall_guess}:

\begin{equation}
    \vec{B'} = \vec{B} + L_b\begin{bmatrix}
  \cos{\theta_B} \\ 
  \sin{\theta_B}
\end{bmatrix}.
\end{equation}

Positive Pivot and Negative Pivot share the same equations, as they both see the entire vine robot body rotating about the pivot point, except in opposite directions. To calculate this out we can start with the location of the point, $\vec{T}$, relative to the contact and pivot points:
\begin{equation}\label{eq:pp_np_setup}
    \begin{split}
    \vec{r}_{T/B}  &= \vec{r}_{T/C}
    \\\vec{B} + L_b\begin{bmatrix}
        \cos{\theta_{B',i}} \\ 
        \sin{\theta_{B',i}}
        \end{bmatrix} &= \vec{C} + L_a\begin{bmatrix}
        \cos(\theta_W + \theta_{C,i}) \\ 
        \sin(\theta_W + \theta_{C,i})
        \end{bmatrix}
    \end{split}
\end{equation}
where \(\theta_{B',i}\) is calculated from rearranging~\Cref{eq:wall_angle}, since it changes with \(\theta_{C,i}\):
\begin{equation}
    \theta_{B',i} = \theta_W - \theta_{C,i} - \theta_T
\end{equation}

If we rearrange~\Cref{eq:pp_np_setup}, we can solve for the point of contact as the robot tip slides and the body pivots:
\begin{equation}\label{eq:lean_wall_guess}
    \vec{C_i} = \vec{B} + L_b\begin{bmatrix}
        \cos{\theta_{B',i}} \\ 
        \sin{\theta_{B',i}}
        \end{bmatrix} - L_{a,i}\begin{bmatrix}
        \cos(\theta_W + \theta_{C,i}) \\ 
        \sin(\theta_W + \theta_{C,i})
        \end{bmatrix}.
\end{equation}

Finally, for Friction-Locked, since this behavior has the vine robot tip stuck at the initial contact point until it changes into another morphology, there is no calculation needed for where the wall is located. However, we can still calculate the evolution of the robot shape and the maximum buckling that will occur at a given contact angle. We begin by noting that the length between the previous pivot point and the turn, \(L_b\), remains constant throughout this event, and this robot segment rotates about the previous pivot point, \(\vec{B}\). This means the new turning joint location, \(\vec{T}'\), lies on the circle centered about \(\vec{B}\) with radius \(L_b\), whose equation is given by:
\begin{equation}\label{eq:squish_circle}
    (T'_{x,i}-B_x)^2 + (T'_{y,i}-B_y)^2 = L_b^2
\end{equation}
where we use the notation $T'_{x,i}$ to indicate the x component of the position vector $\vec{T}'_{i}$, etc.

As the collision angle measurement changes, we can determine where on that circle the turn point lies using the intersection of the line that intersects both the collision point, \(\vec{C_o}\), and the new turning joint location, \(\vec{T}'\), given by:

\begin{equation}\label{eq:squish_line_y}
    T'_{y,i} = (T'_{x,i} - C_{x,o})\tan(\theta_W - \theta_{C,i}) + C_{y,o}
\end{equation}
With these two equations, we can determine the location of the x- and y-components of the turning joint location given the collision angle. For the sake of clarity, we break up the resulting quadratic formula that gives \(T'_{x,i}\):

\begin{equation}\label{eq:squish_line_x}
    T'_{x,i} = \frac{-b\pm\sqrt{b^2-4ac}}{2a}
\end{equation}
where
\begin{equation}
\begin{split}
    a &= 1 + \tan^2(\theta_W-\theta_{C,i}) \\
    b &= 2\tan(\theta_W-\theta_{C,i})\bigl(C_{y,o}-C_{x,o}\tan(\theta_W-\theta_{C,i})\bigr)\\
    & \quad - 2\bigl(B_x-B_y\tan(\theta_W-\theta_C)\bigr) \\
    c &=B_x^2 - L_b^2 + \Bigl(B_y-\bigl(C_{y,o}-C_{x,o}\tan(\theta_W-\theta_{C,i})\bigr)\Bigr) ^2
\end{split}
\end{equation}

This approach will result in two possible \(\vec{T}'\) points from the two intersections between the line and circle. However, we know that the robot will grow over time, so the correct position should be the point that minimizes the distance traveled by \(L_b\) while making sure the total robot length is increasing. If we use our critical angle equations to determine when the robot will shift behavior, we can use that to mark where $\vec{T}'$ will get to before the contact shifts to a new behavior.

Thus, with these equations, we can estimate the movement of a vine robot, turning or not, from a simple set of sensed information. Specifically, we can determine where in space the robot's tip is during contact and the shape of its body by sensing only the vine robot's tip collision angle, total robot length, and the robot starting shape, i.e. initial launch position and angle, and the turn location and angle.

\section{Geometric Simulation for Environment Collisions}\label{sect:simulator}
With a model of the kinematics of vine robots colliding with and passively deforming around obstacles as they continue to grow completed, we now leverage this model to simulate movement through multiple collisions in multi-object environments. This section overviews some simplifying assumptions that arise from the kinematics and discuss the algorithm we use to simulate these vine robot deformations. 

\subsection{Overview of Multi-Contact Modeling}
Vine robot deployments into multi-object environments tend to have much more complex interactions than just a single collision, so we need to understand how these multi-contact deployments build on the kinematics of the robot. Since the kinematics depend primarily on the robot and environment geometry, in particular the previous contact points which act as pivots, each interaction influences the next. While this may initially seem to be a situation where errors compound, we observe some simplifying assumptions which help to stabilize the results of sequential interactions.

\subsubsection{Visibility Graphs as Environment Decompositions}
\begin{figure}[t!]
\centerline{\includegraphics[width=1\columnwidth]{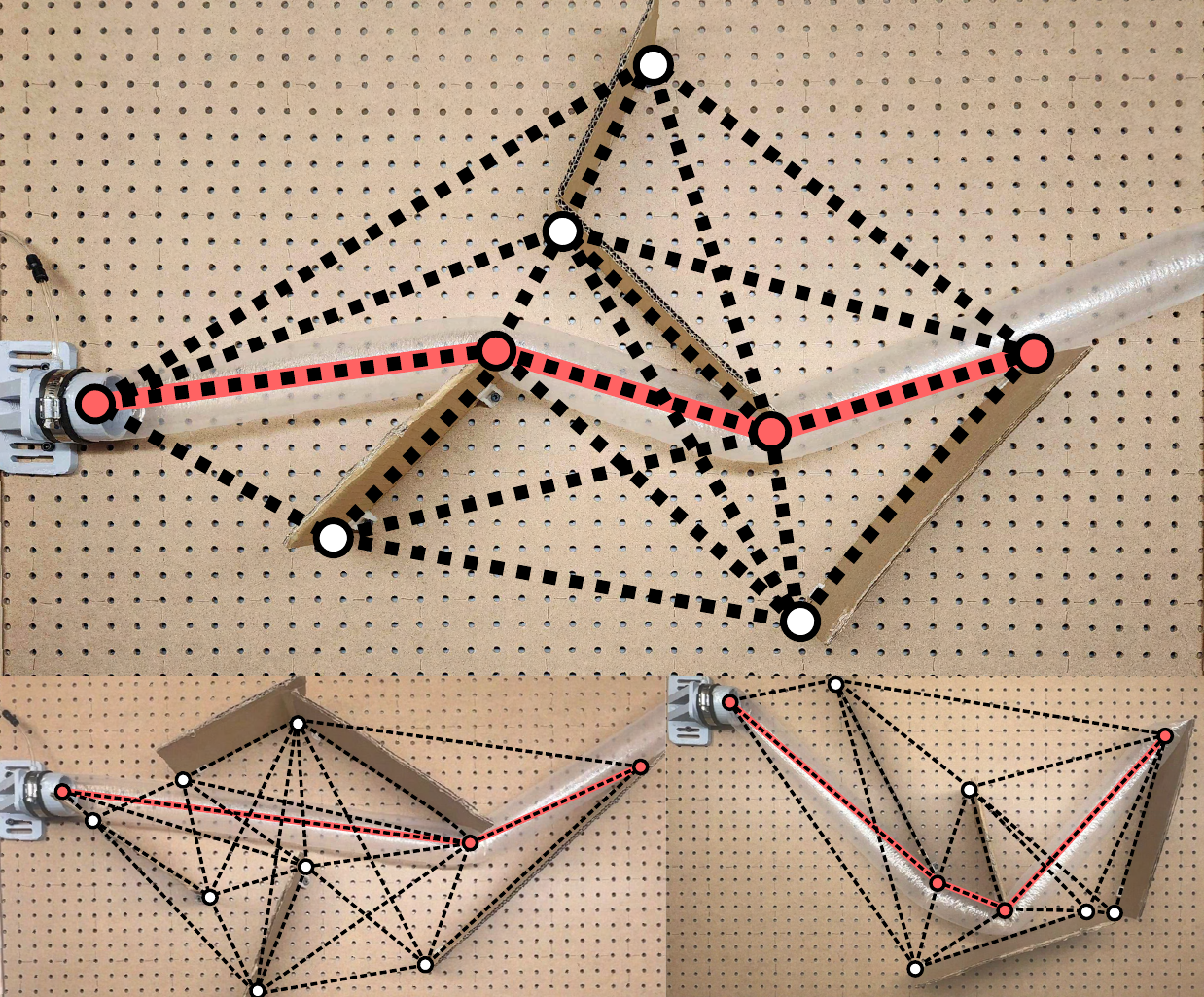}}
\caption{Visibility graphs overlaid onto a manufactured environment with the final shape of the vine robot highlighted in red. Vine robot paths under passive deformation follow paths within the visibility graph.}
\label{fig:VisGraph}
\end{figure}

A critical observation for the multi-contact simulation of vine robots is that all final vine robot shapes which include collision must lie on the visibility graph of the environment plus the starting location of the robot. A visibility graph, in general, contains all obstacle vertices as vertices of the graph and then adds edges wherever two vertices have lines-of-sight on each other, including along obstacle walls. Since the vine robot will slide along a wall until it encounters the end vertex of the wall, obstacle vertices will always act as subsequent potential pivot points. If we repeat this interaction, the shape of the vine robot will sequentially join the start point to a vertex and then obstacle vertices together, therefore, by definition, lying on the visibility graph. Figure~\ref{fig:VisGraph} shows some examples of real vine robot deployments into environments and overlays the visibility graph of the environment plus starting point. We can see that the vine robot's final shape (in red) follows the environment's visibility graph until it leaves the final obstacle interaction. With this observation, simulating a vine robot's movement within an environment can be simplified to 1) calculating the collision behavior based on the initial contact geometry and then 2) identifying the corresponding line of the visibility graph where the robot will go towards (as seen in Figure~\ref{fig:intro}). 

Note, if we strictly consider the centerline of the robot, there will be small differences between the visibility graph and the vine robot's final shape. When the soft body contacts a vertex, the centerline will be offset by the radius of the robot, but when the body wraps around the vertex it deforms to pull the centerline closer to the vertex. Due to this, when translating the environment into simulation, we can reduce the offset by adding a bumper around each obstacle equal to the vine robot radius. This allows us to reduce, but not fully eliminate the offset, though the fact that the robot follows the visibility graph means this error is not likely to stack.

\subsubsection{Vine Robot Obstacle Interactions}\label{sect:VR_Obs_interactions}
In order to make use of the visibility graph as a tool for vine robot modeling, we need to predict which graph edges the vine robot will follow during its navigation. Revisiting the kinematic model in Section~\ref{sect:equations}, a straight vine robot colliding with an object will slide towards the greater angle of collision. Within the context of the visibility graph, the interaction kinematics inform us of whether the robot is rotating clockwise or counter-clockwise, and therefore which line of the visibility graph originating from its pivot it will encounter next. If that line connects to the obstacle currently being contacted, it will be at the end of the wall. Otherwise, it will represent a new contact along the body and a new point to continue the pivoting. An example of this can be seen in the lower right deployment in Figure~\ref{fig:VisGraph}, where the interaction with the bottom wall will be interrupted by contacting the center wall (creating the third pivot point of the robot).

\begin{figure}[t!]
\centerline{\includegraphics[width=.9\columnwidth]{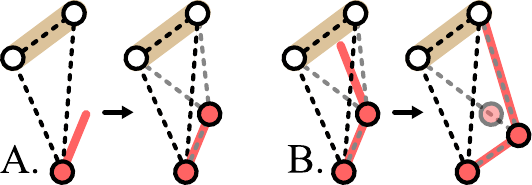}}
\caption{Overview of how turning is integrated into the visibility graph architecture. A) When the vine robot turns, a new vertex with all visible connections is added to the visibility graph. B) As the vine robot grows, its interactions can cause the vertex to be moved and updated in the main visibility graph.}
\label{fig:VG_turning}
\end{figure}

\begin{algorithm}[b!]
\caption{Vine Robot Simulator} 
\label{alg:VR_sim}
\footnotesize
\hspace*{\algorithmicindent} \textbf{Input:} List of polygonal obstacles $O$ with clockwise-ordered vertices, the Visibility Graph $G_V$ generated from $O$, and the vine robot launch conditions: launch point $P_L$, launch angle $\theta_L$, turn location on the robot's length $L_T$, turn angle $\theta_T$, and the robot maximum length $L_{VR}$ \\
\hspace*{\algorithmicindent} \textbf{Output:} Area swept $A$, Walls collided $W$, and the final shape of the vine robot $S$ as a list of vertices.
\begin{algorithmic}[1]

\State Add $P_L$ to $G_V$ and $S$
\State $\vec{n} \gets P_L$, $\theta \gets \theta_L$, \text{turning} $\gets$ false
\While{not $done$}
    \State Calculate remaining robot length $L_\text{rem}$ from $S$
    \State Calculate remaining shape of robot from $\vec{n}$ using $L_\text{rem}$ and $\theta$
    \State Check collision of projected growth against obstacles $O$
    \State Check if turning before collision from $L_T$
    \If{$\neg$collision and $\neg$turning}
        \State Add final position from growth to $S$
        \State \textbf{return}
    \ElsIf{$\neg$turning and $L_T$ before collision}
        \State Calculate location of turning point $\vec{T}$ using $\vec{n}$, $L_T$, and $\theta$
        \State Add $\vec{T}$ to $G_V$
        \State $\vec{n} \gets \vec{T}$, $\theta \gets \theta + \theta_T$, \text{turning} $\gets$ true
        \State \textbf{continue}
    \Else
        \State $\vec{n_d}, \vec{n_h}, \theta_C \gets \texttt{Collision}(\cdots) $\Comment{\Cref{alg:collision}}
        \If{turning}
            \State $\vec{n_d}, \vec{n_h} \gets \texttt{TurningKinematics}(\cdots)$ \Comment{\Cref{alg:turning}}
        \EndIf
    \EndIf
    \State If wrapping around $\vec{n}_{hprev}$ or $\vec{n}$, add to $S$ \Comment{\Cref{sect:vrsim_overview}}
    \State Add line segment $\{ \vec{n_h}, \vec{n_d} \}$ to $W$
    \State Add triangle formed by $\{\vec{n}, \vec{n_h}, \vec{n_d} \}$ to $A$
    \State $\theta \gets atan(\vec{n},\vec{n_d})$% angle between $\vec{n}$ and $\vec{n_d}$
    \State $\vec{n} \gets \vec{n_d}$
\EndWhile
\end{algorithmic}
\end{algorithm}

When the next visibility graph line shows the vine robot reaching the vertex of the obstacle it is contacting, one of three outcomes will occur, determined by the collision angles between the robot orientation and each neighboring edge. The behavior again rises from the robot kinematics, where the vine robot will \textit{only} follow edges where the collision angle is between 90$^o$ and 180$^o$. At a vertex this gives three distinct options: at least one angle is above this range (`skim'), both angles are below this range (`stuck'), or at least one angle is above 90$^o$ and both are below 180$^o$ (`slide'). The `slide' behavior will proceed like a wall contact, moving towards the larger of the valid angles. The `skim' behavior means the robot will leave contact with the obstacle at that point and continue to grow in the direction of its current orientation. Finally, the `stuck' behavior means the robot cannot continue as neither direction will be valid for the kinematics. In general, for convex objects we should only see the `slide' and `skim' behaviors. Only in concave corners will we see a `stuck' interaction. While these describe what happens as the robot reaches the end of a wall, they can also be used to understand the behavior when a collision starts at a vertex. 

\subsubsection{Turns as Modifications of the Visibility Graph}\label{sect:VG_turning}
The addition of turns requires one additional modification of the visibility graph approach. When a discrete-type turn is introduced to the vine robot's movement, we also introduce a vertex to the visibility graph at the location of the turn, as shown in Figure~\ref{fig:VG_turning}A. However, as highlighted previously in Section~\ref{sect:kin_sens}, the location of the turn often moves as the vine robot continues growing (Figure~\ref{fig:VG_turning}B). Only the Pivot-at-Turn morphology keeps this vertex where it is first added.

Unfortunately, we cannot just calculate the final shape and tip position of the vine robot to get the new turn point, as this risks missing contacts along the length as the vine robot deforms. We address this by first calculating the behavior if no extra contacts along the body are made. With the kinematic model, we can predict the behavior, or set of behaviors for the Friction-Locked case, in order to understand which direction the tip and the segment before the turn, $L_b$, move. The following equation determines the expected angle of \(L_b\) given the final vertex we reach without collision:
\begin{equation}\label{eq:turn_loc_update}
    \theta_B = 180 - \tan^{-1}\left(\frac{D_y-B_y}{D_x-B_x}\right)-\sin^{-1}\left(\frac{L_b}{L_c}\sin(\theta_T)\right)
\end{equation}
where \(L_c\) is the distance between the destination vertex, \(\vec{D}\), and the current pivot point, \(\vec{B}\). Note that this equation is derived similarly to \Cref{eq:contact_from_th_B}.

Using the new angle of \(L_b\), we can check that this does not exceed the next visibility graph line from the current pivot. If it does, either segment \(L_b\) or segment \(L_a\) may collide with that point. If that vertex is within distance \(L_b\) from \(\vec{B}\), it will collide with this vertex. If it outside this radius, it may still collide with segment \(L_a\) depending on the final position of that segment. While there are different methods to approach this case, we choose to calculate the polygon formed between the starting and ending shape, and then check whether the vertex is interior to this polygon.

\begin{algorithm}[b!]
\caption{Collision} 
\label{alg:collision}
\footnotesize
\hspace*{\algorithmicindent} %
\textbf{Input:} Current Node $\vec{n}$, Angle $\theta$, and Visibility Graph $G_V$ \\
\hspace*{\algorithmicindent} 
\textbf{Output:} Destination Node $\vec{n_d}$, Hit Node $\vec{n_h}$, Collision Angle $\theta_C$
\begin{algorithmic}[1]

\State $N \gets$ neighborhood of $\vec{n}$ in $G_V$
    \If{$\theta$ is equal to any graph edge angle in $N$} \Comment{Hit vertex}
        \State Hit node $\vec{n_h} \gets$ node from matching edge angle
        % \State $collision\_type \gets vertex$
        \State $\vec{n}_{C1}$, $\vec{n}_{C2}$ $\gets$ adjacent vertices to $\vec{n_h}$
        \State $\theta_{C1}$, $\theta_{C2}$ $\gets$ angles between $\overline{\vec{n}\vec{n_h}}$ and $\overline{\vec{n}\vec{n}_{C1}}$, $\overline{\vec{n}\vec{n}_{C2}}$
        \If{$\max(\theta_{C1}, \theta_{C2}) > 180\degree$} \Comment{Skim}
            \State $\vec{n}_{hprev} \gets \vec{n_h}$
            \State $\theta \gets atan(\vec{n},\vec{n_h})$% angle between $\vec{n}$ and $\vec{n_h}$
            \State $\vec{n} \gets \vec{n_h}$
            \State \textbf{continue} (in~\Cref{alg:VR_sim})
        \ElsIf{$\max(\theta_{C1}, \theta_{C2}) < \theta^*_c$ from \Cref{Eq:straightVR}} \Comment{Stuck}
            \State Add $\vec{n_h}$ to $S$
            \State \textbf{return} (in~\Cref{alg:VR_sim})
        \Else \Comment{Slide}
        \State $\vec{n_d} \gets$ $\vec{n}_{C1}$ if $\theta_{C1} > \theta_{C2}$, else $\vec{n}_{C2}$
        \State $\theta_C \gets \text{max}(\theta_{C1}$, $\theta_{C2})$

        \EndIf
    \Else \Comment{Hit an edge, not a vertex}%edge hit
        \State Project growth and calculate collision point $\vec{n_h}$
        \State $\vec{n}_{C1}$, $\vec{n}_{C2}$ $\gets$ adjacent vertices to $\vec{n_h}$
        \State $\theta_{C1}$, $\theta_{C2}$ $\gets$ angles between $\overline{\vec{n}\vec{n_h}}$ and $\overline{\vec{n}\vec{n}_{C1}}$, $\overline{\vec{n}\vec{n}_{C2}}$
        \If{$\max(\theta_{C1}, \theta_{C2}) < \theta^*_c$ from \Cref{Eq:straightVR}} \Comment{Stuck}
            \State Add $\vec{n_h}$ to $S$
            \State \textbf{return} (in~\Cref{alg:VR_sim})
        \ElsIf{$\theta_{C1} > \theta_{C2}$}
            \State $\theta_C \gets \theta_{C1}$
            \State $\vec{n_B} \gets$ node $\in N$ clockwise from $\theta$
            \State \textbf{if} $\vec{n_B}\equiv \vec{n}_{C1}$ \textbf{then} $\vec{n_d} \gets \vec{n_B}$
            \State \textbf{else} $\vec{n_h} \gets \vec{n_B}$
        \Else
            \State $\theta_C \gets \theta_{C2}$
            \State $\vec{n_B} \gets$ node $\in N$ counter-clockwise from $\theta$
            \State \textbf{if} $\vec{n_B}\equiv \vec{n}_{C2}$ \textbf{then} $\vec{n_d} \gets \vec{n_B}$
            \State \textbf{else} $\vec{n_h} \gets \vec{n_B}$
        \EndIf
    \EndIf

\end{algorithmic}
\end{algorithm}

\subsection{Overview of Vine Robot Simulator}
\label{sect:vrsim_overview}
With the above simplifying assumptions for multi-obstacle environments we can create a simulator that, given a known environment configuration and vine robot deployment parameters, will output the final shape of the vine robot, the contacted walls it collided, and the area its body `swept' through, i.e. moved through, as the robot grew. The simulator takes advantage of the visibility graph to skip intermediate steps that would otherwise be necessary in an iterative approach~\citep{greer2020robust}. The algorithms presented here delineate the principal logic of the simulator used for the experiments in further sections.

Algorithm~\ref{alg:VR_sim} is the high level logic of the simulator, primarily handling modifications to the visibility graph, checking when collisions or pre-formed turns in the robot occur, and updating the final vine robot shape and environment information. Each iteration of the algorithm's main loop calculates the next vertex, if any, of the environment that the robot moves towards and, based on this, modifies the three primary variables of the vine robot: the remaining length, $L_{rem}$, the tip position or current node, $\vec{n}$, and the tip orientation, $\theta$. One concept not explicitly introduced previously, `wrapping' (Line 22), implements the check for when the current or previous encountered vertex acts as a pivot point and should be added to the final robot shape. With few exceptions, this check can be implemented by looking at the previous pivot and checking its line-of-sight on the new destination node $\vec{n_d}$.

Within this main algorithm, two sub-algorithms are used to determine the results of the current collision and returns either a destination node, $\vec{n_d}$, if the robot contacts and slides to the end of a wall, or a hit node, $\vec{n_h}$, if the next vertex is a potential pivot point only. Algorithm~\ref{alg:collision} calculates the next vertex given the straight robot kinematics alone. It first determines if the collision is with an obstacle vertex or an obstacle wall and then determines which vertex the robot ends at next. In the first case, that vertex is evaluated as a `skim', `stuck', or `slide' point to determine whether it or one of its adjacent vertices should be added next. In the second case, if the contact is near perpendicular, we consider the robot stuck, otherwise we find the next node in the visibility graph for the robot to pivot towards. If there was no turn proceeding the collision, we check whether the new hit or destination node causes wrapping at a previous pivot and we update the wall collisions and swept area. If there is turning, Algorithm~\ref{alg:collision}'s outputs are first fed into Algorithm~\ref{alg:turning}. The straight robot kinematics give a good approximation to the turning kinematics, so the algorithm uses that as a start point to determine how the collision is altered, following~\Cref{Eq:static_squish,Eq:squish_lean,Eq:lean_static_alt} to determine collision behavior and using \Cref{eq:squish_line_x,eq:squish_line_y} if needed as in the Friction-Locked case.

The simulator is computationally efficient due to the visibility graph representation. Collisions can be computed directly from the next vertex in the visibility graph, resulting in processing $O(V)$ collision events per deployment where $V$ is the number of visibility graph vertices. The visibility graph itself is constructed once per environment in $O(n^2 \log n)$ time for $n$ obstacle vertices using a standard algorithm.

%%%

\begin{algorithm}[b!]
\caption{Turning Kinematics} 
\label{alg:turning}
\footnotesize
\hspace*{\algorithmicindent}
\textbf{Input:} Destination Node $\vec{n_d}$, Hit Node $\vec{n_h}$, Collision Angle $\theta_C$, Turning Point $\vec{T}$ \\
\hspace*{\algorithmicindent} 
\textbf{Output:} Destination Node $\vec{n_d}$, Hit Node $\vec{n_h}$
\begin{algorithmic}[1]

    \State $L_a \gets ||\vec{T}-\vec{n_h}||$, $L_b \gets ||\vec{T}-\vec{n}||$

    \State Calculate $\theta_{\text{PP}|\text{PaT}}, \theta_{\text{PaT}|\text{FL}}, \theta_{\text{FL}|\text{NP}}$ from \Cref{Eq:lean_static_alt,Eq:static_squish,Eq:squish_lean}
    \If{$\theta_C \leq  \theta_{\text{PP}|\text{PaT}}$ or $\theta_C \geq  \theta_{\text{FL}|\text{NP}}$} \Comment{Positive or Negative Pivot}
        \State Find body position $\theta_B$ using \Cref{eq:turn_loc_update}
        \State Find new turn location $\vec{T}'$ by extending from $\vec{n}$ using $\theta_B$ and $L_b$
        \State Check if robot collides while sliding towards $\vec{n_d}$ \Comment{\Cref{sect:VG_turning}}
        \If{collision}
            \State Update $\vec{n}$ and $\theta$ with collided point
            \State Add collided point to $S$
            \If{collided point is after $\vec{T}$}
                \State turning $\gets$ false
            \Else
                \State Update $\vec{T}'$ again by extending from $\vec{n}$ using $\theta_B$ and $L_b$
            \EndIf
            \State \textbf{continue} (in~\Cref{alg:VR_sim})
        \Else{}
            \State $\vec{n} \gets \vec{n_d}$,
            $\theta \gets \theta_B + \theta_T$
        \EndIf
        % \State Update $AS$ using $T$, $T'$, $n_h$, and $\vec{n}$
        
    \ElsIf{$\theta_C \leq  \theta_{\text{PaT}|\text{FL}}$} \Comment{Pivot-at-Turn}
        \State turning $\gets$ false%yeah, this is the easiest one

    \ElsIf{$\theta_C <  \theta_{\text{FL}|\text{NP}}$} \Comment{Friction-Locked}
        \State $\theta^* \gets \argmin_{\alpha \in \{\theta_{PaT|FL}, \theta_{FL|NP}\}} |\alpha - \theta_C|$
        \State Find $\vec{T}'$ using $\theta^*$ and \Cref{eq:squish_line_x,eq:squish_line_y} (replace $\theta_C$ with $\theta^*$)
        \If{$\vec{T}'$ is real}
         \textbf{continue} (in~\Cref{alg:VR_sim})
        \Else~turning $\gets$ false
        \EndIf
    \EndIf
    \State Add swept area between $\{\vec{n}, \vec{T}, \vec{n_h}\}$ and $\{\vec{n}, \vec{T}', \vec{n_d}\}$ to $A$

\end{algorithmic}
\end{algorithm}

\subsection{Validation of the Simulator}
\begin{figure}[t!]
\centerline{\includegraphics[width=1\columnwidth]{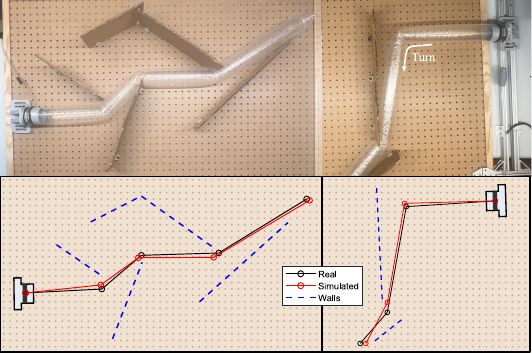}}
\caption{Deployments in a real environment (top row) and the comparison of the extracted shape to the same deployments using the simulator (bottom row). The blue dashed lines show the extracted walls from the real environment, the black line shows the centerline of the real robot, and the red line shows the centerline of the simulated robot. }
\label{fig:simvreal}
\vspace{-1.0em}
\end{figure}

In order to verify the use of the vine robot multi-contact simulator, we explored its accuracy in predicting vine robot movement through real environments. We constructed several environments and deployed vine robots into them at a variety of starting positions and angles. The vine robot was constructed from low-density polyethylene tubing with a nominal radius of 3.23~cm and a length of approximately 1~m. The robot's motions were recorded using an overhead camera (Figure~\ref{fig:simvreal}), and the videos were analyzed to extract contacted walls, area swept, and the final vine robot position. We then simulated the same environments (with a buffer added for the vine robot radius) and the vine robot conditions using Algorithm~\ref{alg:VR_sim}. Figure~\ref{fig:simvreal} shows examples of both the real deployment (top row) and the simulated deployment (bottom row).

With these shapes, we can now compute how accurately the simulator predicted the path taken by the vine robot. We extract the vertices that define the shapes and calculate the distance between the actual and simulated positions for those vertices. Specifically, we calculate the euclidean distance of matching points in the shapes, excluding the starting point. When calculated across all the experiments, we see a mean error of 3.40~cm, which is on the same order of magnitude as the robot radius, 3.23~cm.

\section{Passive Deformations for Mapping Unknown Environments}\label{sect:experiments}
With the kinematic model and method to simulate deployments established, we can finally demonstrate the use of the vine robot passive deformation as a tool to map an unknown environment. In this section we explore an unknown environment through sequentially-informed vine robot deployments using a Monte Carlo-inspired algorithm to estimate the robot deployment which will generate the highest reward, i.e. the most new information. This approach is tested in environments with uniformly and non-uniformly shaped obstacles and both with and without turning. Finally, this approach is tested in a real-life demonstration to estimate an unknown environment purely from sensing the vine robot deformations under contact, demonstrating environment mapping without previous information.

\subsection{Problem Formulation}
In previous work, we showed that it was possible to achieve significant mapping in relatively cluttered environment \textbf{\textit{if}} the ideal combination of vine robot deployments could be selected~\citep{Fuentes2023}. Of course, this perfect selection can only be achieved by already having full knowledge of the environment, so to map without any initial prior knowledge, we need an approach that integrates the collected information from all previous deployments to make educated predictions on what next vine robot to deploy. Ideally, each newly selected deployment will collect new information and over time will approach the best selection at that point in time.

\subsubsection{Vine Robot Action Space}
The vine robot action space for both the simulated and physical deployments are comprised of four variables: launch location, launch angle, turning length, and turning angle. This action space is discretized in each of these variables.

\begin{figure}[t!]
\centerline{\includegraphics[width=0.95\columnwidth]{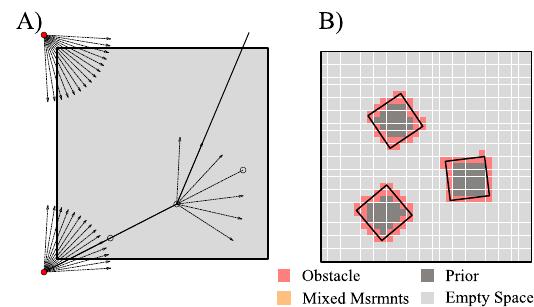}}
\caption{A) Distribution of the variables used to create the full action space of 570 deployments, showing one example, and B) the rubric used to score deployments and beliefs.}
\label{fig:actions_and_rubric}
\vspace{-1.0em}
\end{figure}

In many real exploration or mapping tasks, there are limited entry options into the unknown environment. Thus, in this work, we improve on our previous paper's assumptions~\citep{Fuentes2023}, by limiting the starting positions to only two, both on the same side of the environment. For launch angle, we use a uniform distribution of discrete angles. For the turn variables, it can start at one of a few discrete turning lengths and then turn the robot in a uniform spread of angles centered around 0, which is equivalent to no turning. Figure~\ref{fig:actions_and_rubric}A shows a selection of each variable within these actions. In total for the tests with turning there are 570 actions.

\begin{figure*}[t!]
\centerline{\includegraphics[width=1.9\columnwidth]{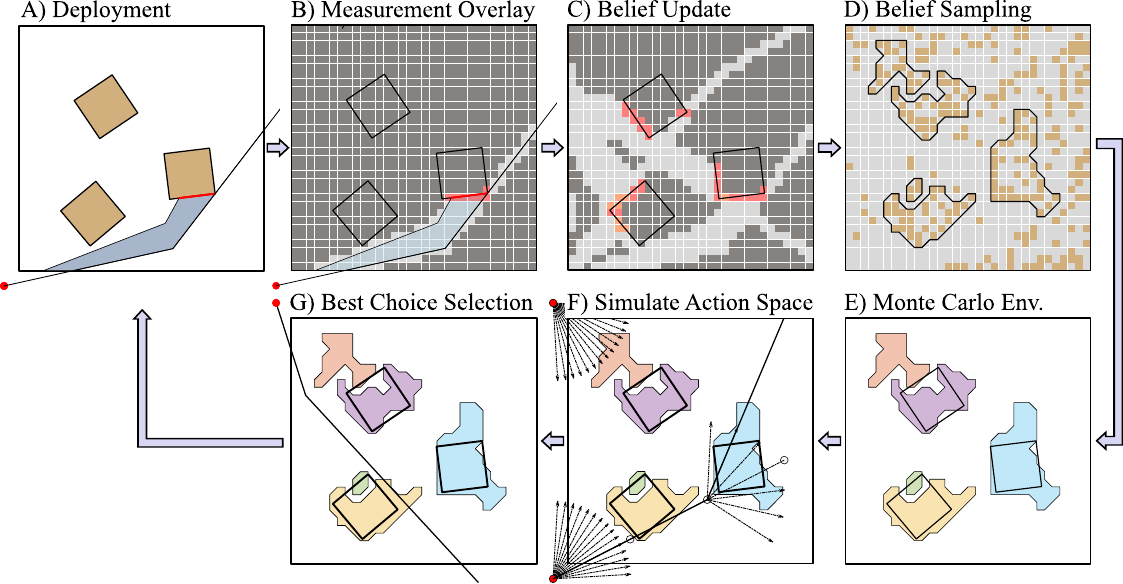}}
\caption{Example of a single loop of the algorithm. First, A) the vine robot is deployed and B) its measurements are used to create the individual belief, which then C) updates the cumulative belief. From this, D) the belief is sampled to create potential environments and cells are grouped to E) select obstacles for Monte Carlo (MC) environments (only one example is pictured). F) The entire action space is simulated against each MC environment. Finally, G) the best action across all simulated environments is then chosen to be deployed into the ``real" environment, starting the loop anew.}
\label{fig:one_loop}
\vspace{-1.0em}
\end{figure*}

\subsubsection{Environment Grid Decomposition}
In order to represent the sensed information about the environment in a computationally efficient way, and readily sample potential environments given the current information, we use a discrete space decomposition. In this work, a square grid decomposition was implemented. To ensure this decomposition remains consistent across all tested environments, each environment was generated to be within the same square boundary and decomposed into a 35 by 35 grid represented by a matrix. The value for the square within the matrix, ranging from 0 to 1, represents the likelihood that square is occupied by an obstacle, with 0 meaning it is empty and 1 meaning it is partially or fully occupied by an obstacle. The values in the squares start at a prior of $\frac{1}{3}$ for the work in this paper, representing an assumption of mostly empty space. During a mapping task we will have two main types of spatial decompositions: ``rubrics" which represent the decomposition of the real or sampled environment and ``beliefs" which represent the information the current and/or previous deployments have found. Belief matrices elements start equal to the prior. Figure~\ref{fig:actions_and_rubric}B shows an example rubric.

\subsubsection{Updating Environment Belief from Deployment Measurements}
When a vine robot is deployed into an environment, it will gather information about areas in the environment that are occupied or unoccupied. An example is shown in Figure~\ref{fig:one_loop}A, where the swept area (blue) and line of the final robot path (black line) will be unoccupied and the wall collision (red) will be occupied. By overlaying these shapes onto the grid decomposition, two separate matrices can be generated, the hit matrix, \(\mathbf{H}\), which carries occupied information, and the miss matrix, \(\mathbf{M}\), which carries unoccupied information. This involves marking all grid cells which are partially or fully covered by the sensed information. From these matrices, the combined belief matrix, \(\mathbf{B}\), can be created element-wise:
\begin{equation}\label{eq:belief}
    \mathbf{B}_{i,j} = p\Bigl(1 - \texttt{bitor}(\mathbf{H}_{i,j}, \mathbf{M}_{i,j})\Bigr) + \frac{\mathbf{H}_{i,j}}{\mathbf{H}_{i,j}+\mathbf{M}_{i,j}}
\end{equation}
where \(p\) is the prior value. Note, we only calculate the belief this way when there is either a hit or a miss, otherwise the value is left at the prior. Figure~\ref{fig:one_loop}B shows an example of a single deployment's belief.
To update the overall belief from a deployment, we first compute cumulative hit and miss matrices by performing a \texttt{bitor} operation between the previous matrices and the newly acquired one. We then use~\Cref{eq:belief} to create the cumulative belief (Figure~\ref{fig:one_loop}C).

\begin{figure*}
    \centerline{\includegraphics[width=1\linewidth]{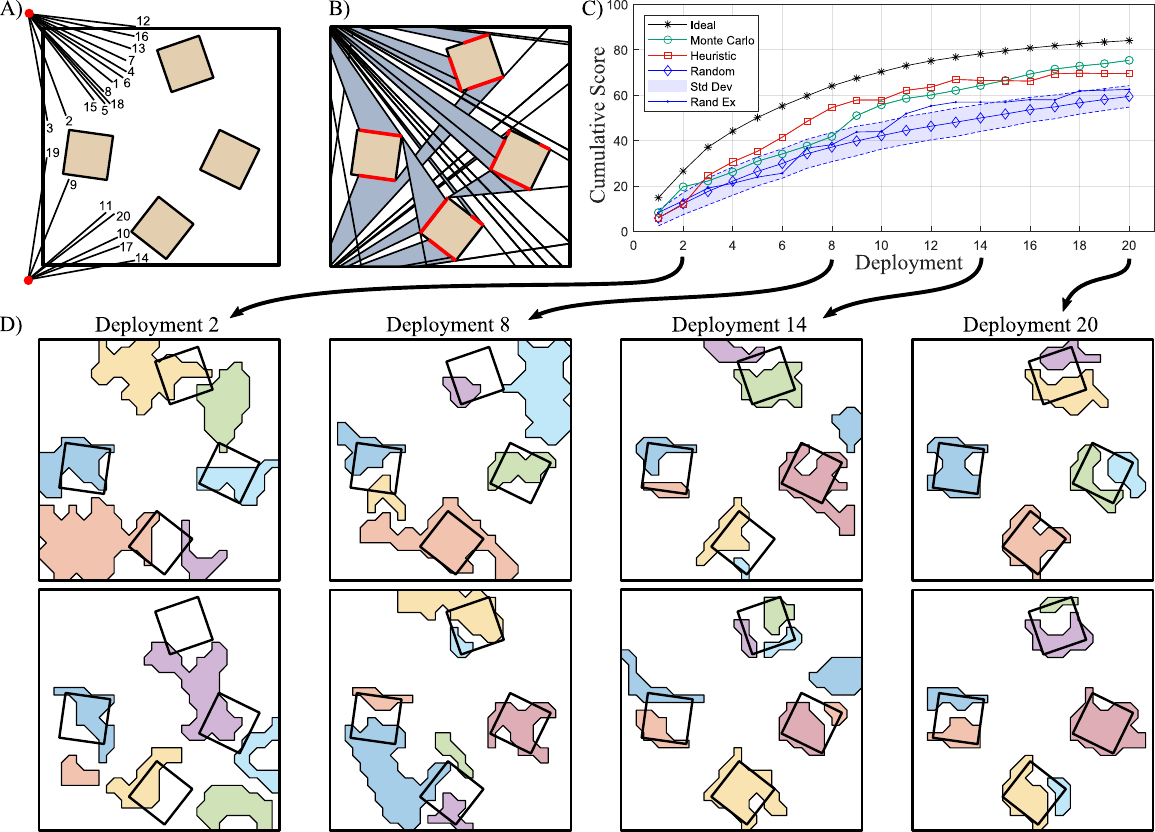}}
    \caption{An expanded example of results for a uniform environment using an action space with no turns. A) Selected deployment actions before launching and B) the same selection of deployments, after launch, including the \textit{new} information gathered. Information from earlier deployments is visually ``on top of" that from later deployments. C) Performance comparison of our proposed algorithm compared to ideal selections and purely random ones. D) Examples of the estimations of the environments after 2, 8, 14, and 20 deployments showing improvement in environment estimation.}
    \label{fig:ex_of_ex}
    \vspace{-1.0em}
\end{figure*}

\begin{figure*}[t!]
\centerline{\includegraphics[width=2\columnwidth]{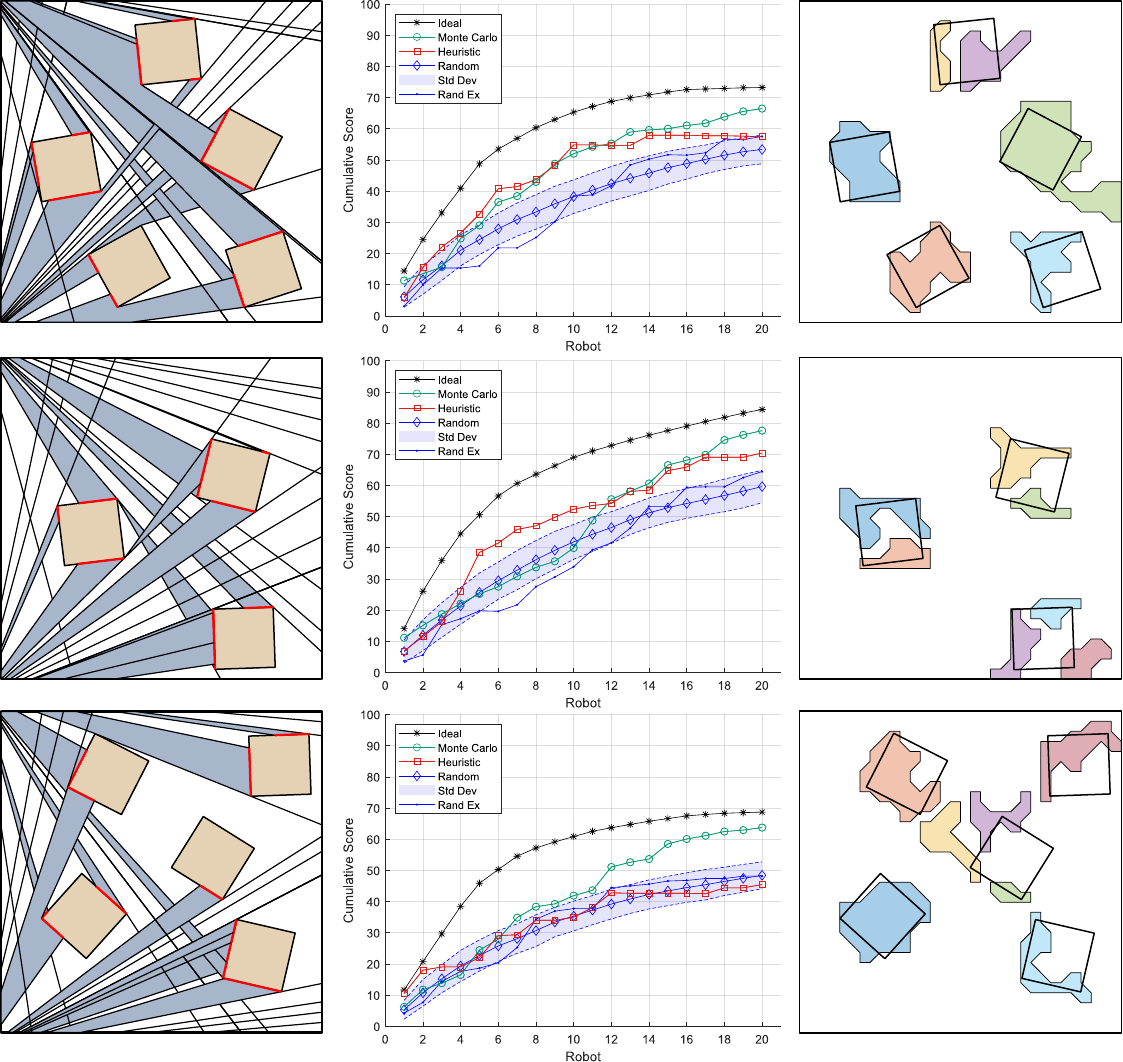}}
\caption{Example results of MC algorithm on \textit{uniform} environments with \textit{straight} vine robots. (Left) Gathered information from 20 vine robots. (Center) Cumulative score compared to Ideal, Random, and Heuristic. (Right) Example MC environment after final loop.}
\label{fig:sqr_str_ex}
\vspace{-1.0em}
\end{figure*}

\subsubsection{Comparing Deployments and Environments through Scoring}
To score a belief, \(\mathbf{B}\), against an environment and evaluate the gained information, we calculate a weighted score, \(S\), for that belief, using:
\begin{equation}\label{eq:scoring}
    S = 100\frac{S_\mathbf{B}-S_\mathbf{P}}{100-S_\mathbf{P}}
\end{equation}
where \(S_\mathbf{P}\) is the unweighted score of the prior matrix, \(\mathbf{P}\), and \(S_\mathbf{B}\) is the unweighted score of the belief. To calculate an unweighted score \(S_\mathbf{X}\) for some matrix \(\mathbf{X}\), we use:
\begin{equation}
    S_\mathbf{X} = 100\ \overline{\ \mathbf{1}- |\mathbf{X - \mathbf{R}}|\ }
\end{equation}
where \(\mathbf{R}\) is the rubric matrix, i.e. the true occupancy matrix, and \(\mathbf{1}\) is a matrix of all ones. These equations allow us to compare the success of deployments and beliefs against the prior matrix, which would score a 0, and the rubric, which would score 100.

\subsubsection{Monte Carlo Simulation}
We use the Monte Carlo (MC) method to sample possible environments given the current belief matrix and then calculate the expected reward for each element of the action space. To generate an environment, for each cell we sample from a uniform distribution between 0 and 1; if the sampled number is below the belief value, the cell will be marked as occupied, otherwise it will be unoccupied. An example of one such sampled matrix is shown in Figure~\ref{fig:one_loop}D. This is likely to result in small scattered obstacles, so we group batches of neighboring diagonal and orthogonal elements containing obstacles and select the largest 6-8 groups, making sure all cells in the current hit grid are contained in obstacles. Finally, a convex hull of each of group is created to form the boundaries of the generated obstacles for a smoother polygonal shape. Our example MC environment is shown in Figure~\ref{fig:one_loop}E. A total of five MC environments are generated at each loop and the entire action space of robots is deployed in each environment (Figure~\ref{fig:one_loop}F). We then select the action with the best combined score among the entire space, as calculated by~\Cref{eq:scoring} before sending the action into the real environment (Figure~\ref{fig:one_loop}G), thus restarting the loop.

The computational cost of each planning loop scales with the number of Monte Carlo samples, the action space size, and grid resolution. For $k$ sampled environments (we use $k=5$) and action spaces of $\vert A \vert$ (that is, 100 for no turning or 570 with turning), each loop evaluates $k\vert A\vert$ simulations; in our experiments, this ranges from $500$ to $2850$, which can be ran in parallel. The grid resolution affects the sampling of environments, as cells are sampled and clustered into convex obstacles in $O(n^2)$ time for a $n\times n$ grid. For larger environments, adaptive strategies could reduce the action space or sample count while preserving decision quality through importance sampling or coarse-to-fine search, however this is left for future work.

\subsection{Simulation Results}
To explore how well passive vine robot deformations can map an unknown environment in our approach, we test the approach varying environment and action space composition. In order to contextualize the performance of our proposed MC-inspired algorithm, we compare it to three other approaches for action selection: Ideal, Random, and Heuristic. The Ideal approach selects the best action from the action space at each step, using full knowledge of the environment, to represent an upper bound on the score, similar to our previous work~\citep{Fuentes2023}. The Random approach selects and launches a random set of deployments from the action space and compiles the score, repeating this random sampling 100 times to determine the average and standard deviation expected to be achieved by a random selection policy. Finally, the Heuristic approach applies a more intuitive search method based on where the largest areas of uncertainty remain. When a robot is deployed and reaches the other side of the boundary space, it will split the unknown region of the environment with the information of its body. With these various slices of the bounded area, the largest unknown area is then selected and its center aimed at for the next deployment, effectively always attempting to target the sub-region of the environment with the least information.

Figure~\ref{fig:ex_of_ex} shows an example mapping with uniform square obstacles and straight robot action space. A total of 20 actions are selected as shown in Figure~\ref{fig:ex_of_ex}A-B. The cumulative score of the actions are shown in Figure~\ref{fig:ex_of_ex}C compared to Ideal, Random, and Heuristic selections and example MC environments at a sequence of loops are shown in Figure~\ref{fig:ex_of_ex}D. A consistent observation across the experiment is the rise of accuracy in estimating the environment, shown both quantitatively through the scoring, and qualitatively through the generated MC environments. While early loops show scores within the range of Random selection and significant variation in the generated environments, a significant improvement is seen around loop 9, and the sets of generated environments in loops 14 and 20 begin to be more consistent and accurately approach the true environment.

Similar performance was seen across a range of environments, with Table~\ref{tb:all_results} showing the average (and standard deviations) of the final scores across the two major variables, obstacle uniformity and action spaces with or without turning.

\begin{table}[t!]
\centering
\begin{tabular}{c|cc|cc|}
\cline{2-5}
                                    & \multicolumn{2}{c|}{Uniform}  & \multicolumn{2}{c|}{Non-Uniform} \\ \cline{2-5} 
                                    & \multicolumn{1}{c|}{Straight}   & Turning   & \multicolumn{1}{c|}{Straight}   & Turning   \\ \hline
\multicolumn{1}{|c|}{Ideal}         & \multicolumn{1}{c|}{\begin{tabular}[c]{@{}c@{}}73.37\\  $\pm$ 6.41\end{tabular}} & \begin{tabular}[c]{@{}c@{}}86.05\\  $\pm$ 2.88\end{tabular} & \multicolumn{1}{c|}{\begin{tabular}[c]{@{}c@{}}49.01\\ $\pm$11.92\end{tabular}} & \begin{tabular}[c]{@{}c@{}}72.81\\ $\pm$5.88\end{tabular} \\ \hline
\multicolumn{1}{|c|}{Monte Carlo} & \multicolumn{1}{c|}{\begin{tabular}[c]{@{}c@{}}64.09\\ $\pm$ 7.73\end{tabular}} & \begin{tabular}[c]{@{}c@{}}70.53\\ $\pm$ 3.82\end{tabular} & \multicolumn{1}{c|}{\begin{tabular}[c]{@{}c@{}}40.85\\ $\pm$ 11.33\end{tabular}} & \begin{tabular}[c]{@{}c@{}}56.55\\ $\pm$ 8.23\end{tabular} \\ \hline
\multicolumn{1}{|c|}{Heuristic} & \multicolumn{1}{c|}{\begin{tabular}[c]{@{}c@{}}52.65\\ $\pm$ 9.93\end{tabular}} & \begin{tabular}[c]{@{}c@{}}56.77\\ $\pm$ 5.64\end{tabular} & \multicolumn{1}{c|}{\begin{tabular}[c]{@{}c@{}}36.86\\ $\pm$ 9.64\end{tabular}} & \begin{tabular}[c]{@{}c@{}}45.16\\ $\pm$ 9.16\end{tabular} \\ \hline
\multicolumn{1}{|c|}{Random} & \multicolumn{1}{c|}{\begin{tabular}[c]{@{}c@{}}50.42\\ $\pm$ 4.62\end{tabular}} & \begin{tabular}[c]{@{}c@{}}55.09\\ $\pm$ 1.81\end{tabular} & \multicolumn{1}{c|}{\begin{tabular}[c]{@{}c@{}}36.37\\ $\pm$ 7.88\end{tabular}}  & \begin{tabular}[c]{@{}c@{}}43.77\\ $\pm$ 6.32\end{tabular} \\ \hline

\end{tabular}\caption{Summary of simulation results across all types of environments and action spaces tested.}
    \label{tb:all_results}
\vspace{-1.0em}
\end{table}

\subsubsection{Performance in Environments with Uniform Objects}
The first set of tests were done on environments comprised of 3-5 randomly placed uniform squares within the boundary of the environment, and an action space of 50 straight robot launch angles at each starting position, totaling 100 unique actions. Figure~\ref{fig:sqr_str_ex} shows a selection of tested environments and the results, with the first column showing the environments with the selected vine robot deployments, the second column showing the MC-algorithm performance against Ideal, Random, and Heuristic selections, and the final column showing an example MC environment generated after the final loop. Overall the performance is well above random selection, but without turns, there are some areas of the map that cannot be accessed, as can be seen in the top and bottom rows. Additionally, in less cluttered environments, like the middle row, it takes longer to perform better than Random and Heuristic as many actions do not hit obstacles. Across most of the environments, the Heuristic method tends to start strong, then either performs comparably to MC deployments or comparably to Random ones. However, across all environments tested, it always loses out to the MC performance.

\begin{table}[t!]
    \centering
    \begin{tabular}{|c|c|c|}
        \hline
         Info Type & Avg Score & Std Dev\\ \hline
         Area Only & 51.22 & 7.18 \\ \hline
         Wall Only & 7.10 & 2.46 \\ \hline
         Wall + Area & 58.32 & - \\ \hline
         Nominal & 64.09 & 7.73\\
         \hline
    \end{tabular}\caption{Overall results of the uniform environments with limited information gain in comparison to their combined information and nominal.}
    \label{tb:split_info}
    \vspace{-1.0em}
\end{table}

\begin{figure*}[t!]
\centerline{\includegraphics[width=2\columnwidth]{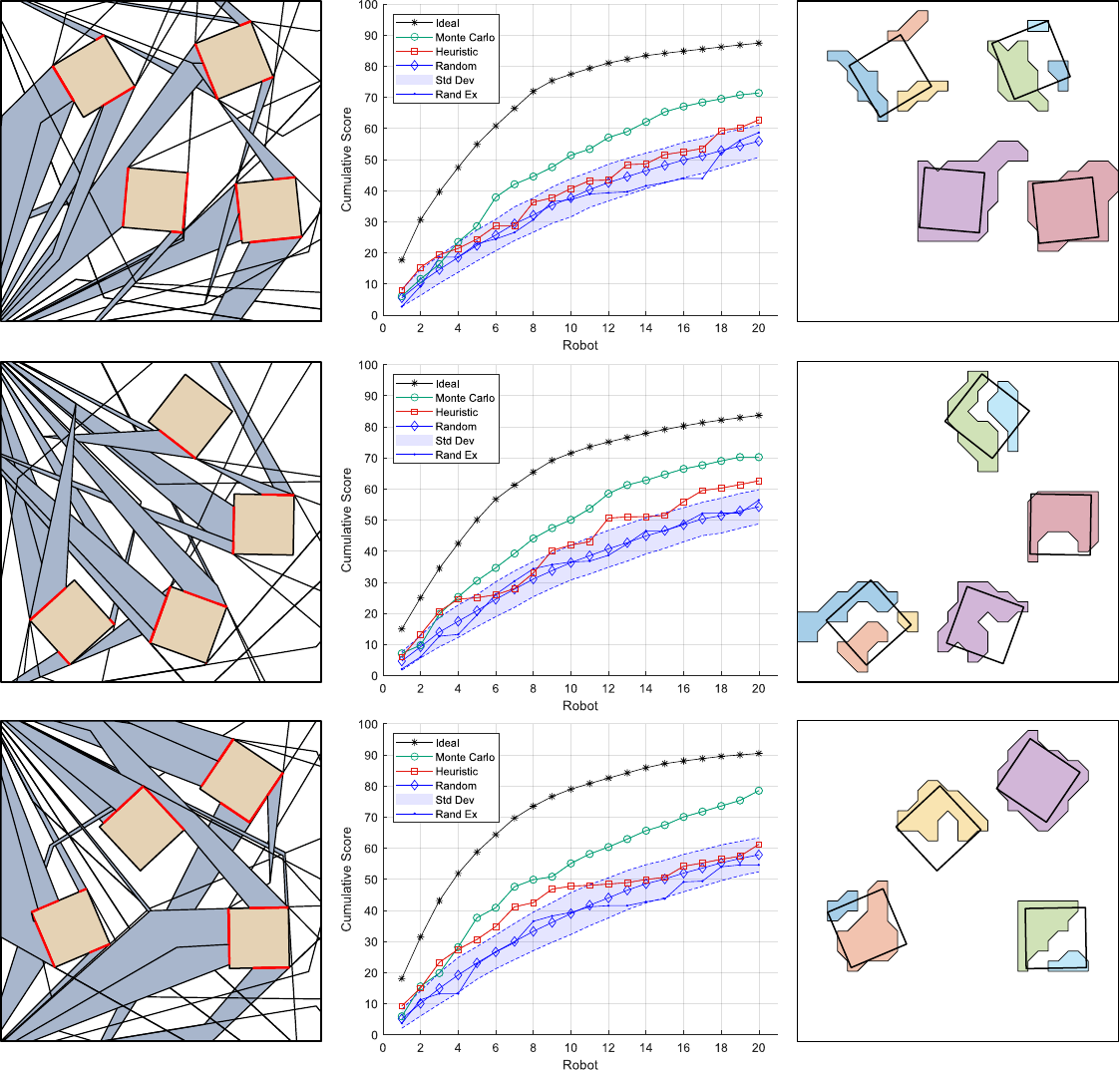}}
\caption{Example results of MC algorithm on \textit{uniform} environments with vine robots with \textit{turns}. (Left) Gathered information from 20 vine robots. (Center) Cumulative score compared to Ideal, Random, and Heuristic. (Right) Example MC environment after final loop.}
\label{fig:sqr_trn_ex}
\vspace{-1.0em}
\end{figure*}

\subsubsection{Limitation on Sensed Information.}
Throughout this work, it has been consistently assumed that having both swept area and wall collision data, working together, is critical to informed environment exploration, motivating the kinematic modeling. To explore this assumption, the uniform environments were retested while only providing half the available information, either swept area information or wall collision information. The results of this exploration compared to the approach with full information are shown in Table~\ref{tb:split_info}. We also include the sum of the area-only and wall-only scores.

The results show how the majority of the score is generated by the unoccupied area. However, this score is not consistently above the Random or Heuristic selection when all their information is given. When both halves of performances are reunited, their summed success is still well below the original approach, though still outside the Random selection range and comparable to Heuristic's success. However, different actions were selected by area-only and wall-only runs, so this information represents, on average, 36 deployments, suggesting the nominal approach is able to choose actions which prioritize both area and wall information.

\begin{figure*}[t!]
\centerline{\includegraphics[width=2\columnwidth]{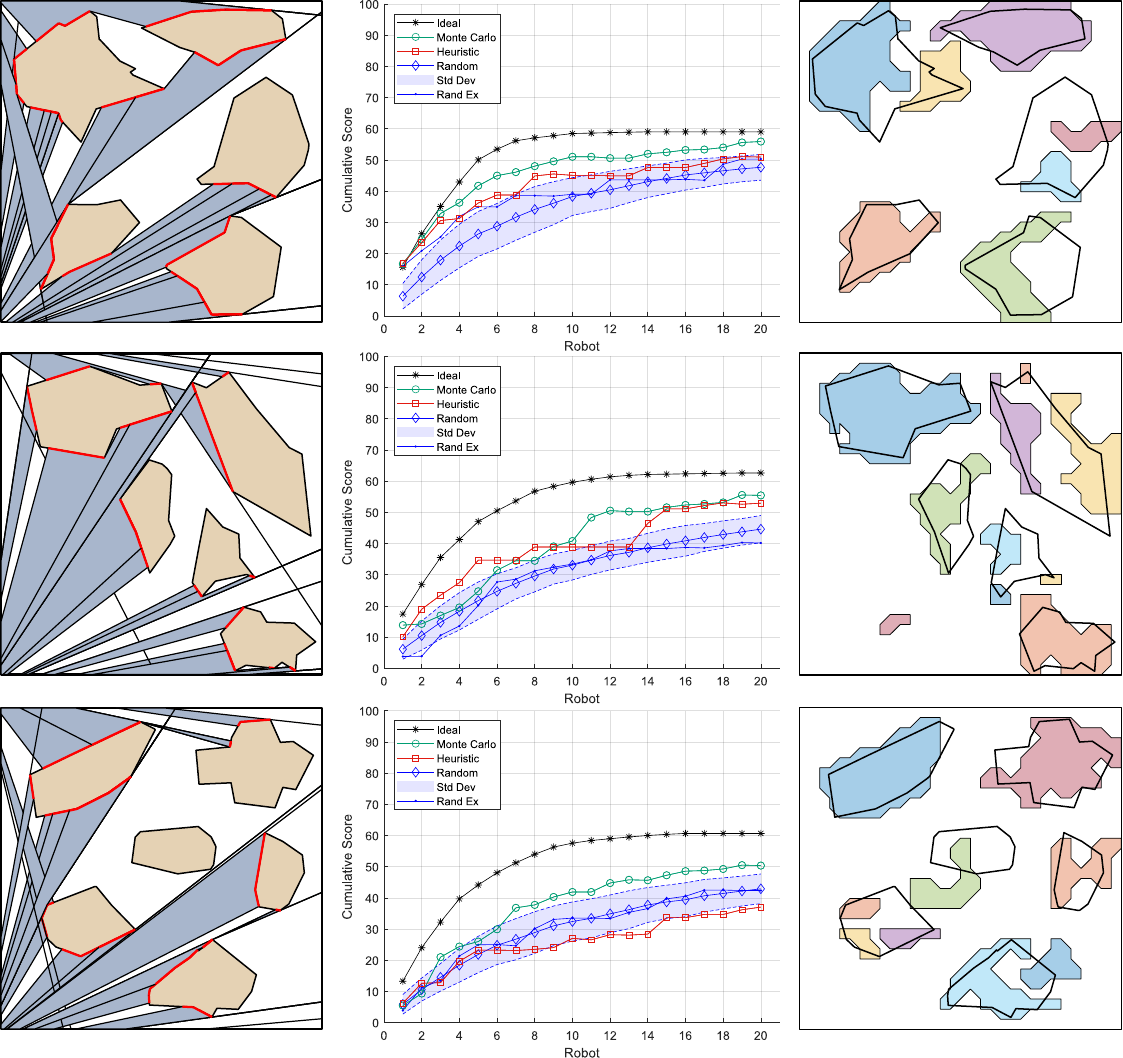}}
\caption{Example results of MC algorithm on \textit{non-uniform} environments with \textit{straight} vine robots. (Left) Gathered information from 20 vine robots. (Center) Cumulative score compared to Ideal, Random, and Heuristic. (Right) Example MC environment after final loop.}
\label{fig:ABC_str_ex}
\vspace{-1.0em}
\end{figure*}

\subsubsection{Effect of Turning in the Action Space}
The first increase in complexity is to introduce turns to the action space, which are expected to help reach spaces the robot would not otherwise have been able to. Due to the significant increase in the size of the action space by adding two more variables, we select a smaller range of types of turns and do not include actions which are unlikely to generate any information. As well, to increase the diversity of selectable turn actions, we reduce the range of deployment angles from 50 to 15, while still maintaining the same two deployment positions. The options for the turning locations were at 20\%, 40\%, and 60\% of the total length of the vine robot, and the turning angles that were implemented were -60, -40, -20, 0, 20, 40, and 60 degrees. This creates an action space of 570 unique deployments, compared to the original 100. The same uniform environments were tested and the increased scores are reflected in Table~\ref{tb:all_results}. The examples in Figure~\ref{fig:sqr_trn_ex} show that there are no longer unreachable areas of the environment and the MC-algorithm selection outpaces Random and Heuristic selections after only a few deployments. Despite the vastly increased action space, our proposed method is able to select useful actions far more effectively.

\begin{figure*}[h!]
\centerline{\includegraphics[width=2\columnwidth]{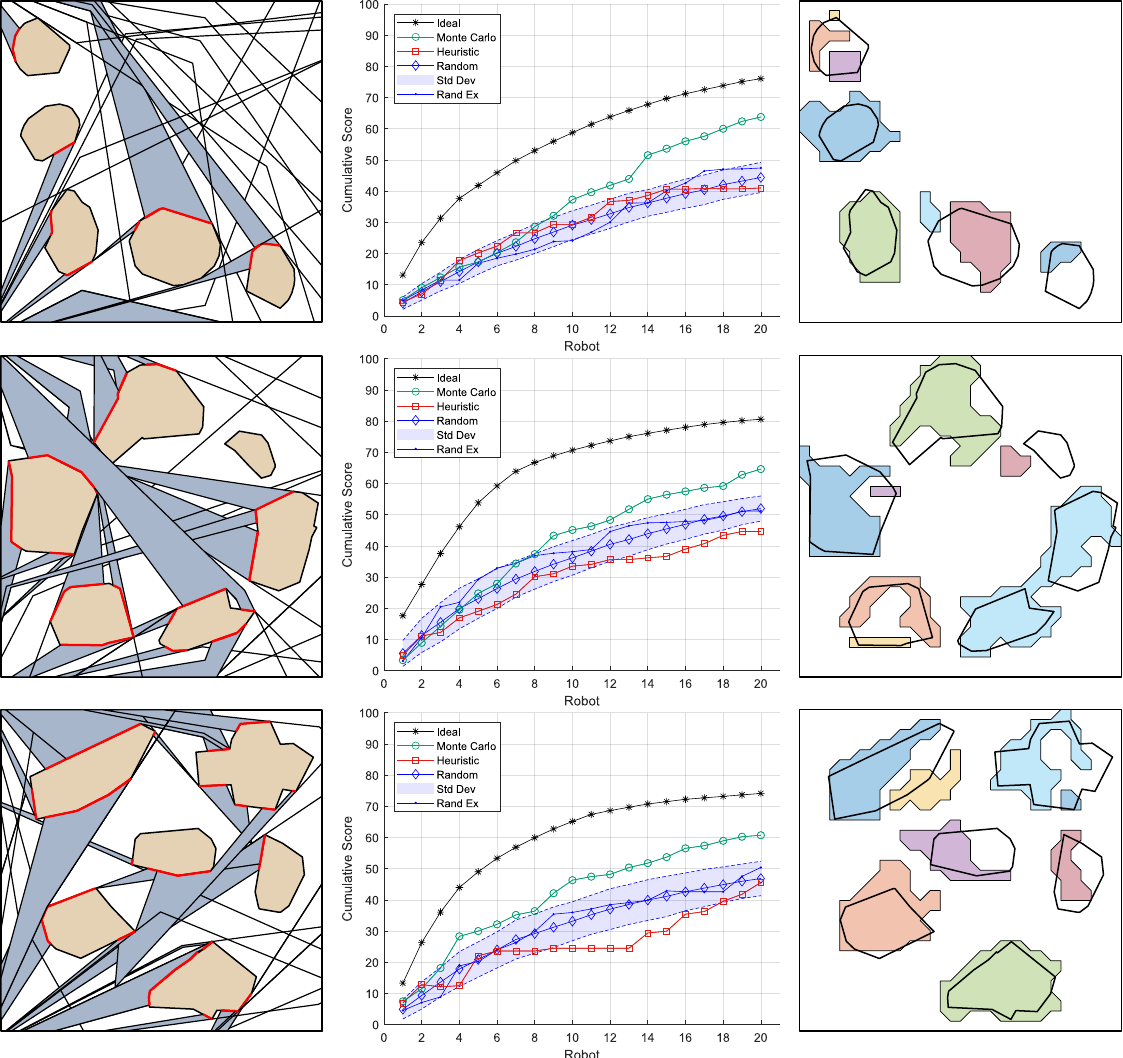}}
\caption{Example results of MC algorithm on \textit{non-uniform} environments with vine robots with \textit{turns}. (Left) Gathered information from 20 vine robots. (Center) Cumulative score compared to Ideal, Random, and Heuristic. (Right) Example MC environment after final loop.}
\label{fig:ABC_trn_ex}
\vspace{-1.0em}
\end{figure*}

\subsubsection{Performance in Environments with Non-Uniform Objects}
Finally, we test the algorithm with non-uniform obstacles, including obstacles with concavity. To create random obstacles, the convex hull of a random splatter of points is used, which is then scaled and placed initially randomly in the environment boundary before being manually adjusted to avoid obstacles being positioned closer than a single grid cell apart. This is repeated until 3-6 random obstacles have been randomly placed inside the bounding region. Note, this method of generating environments results in the obstacles taking up more space in the boundary than the squares, so scores may not be completely comparable to square environments, as previous work determined obstacle density has an effect on the success of passive vine robot mapping~\citep{Fuentes2023}. Eleven total non-uniform environments were tested, with the number and size of obstacles intentionally varied to create a range of unique environments for the simulated vine robots to navigate. A few examples of these experiments are shown in Figures~\ref{fig:ABC_str_ex} and~\ref{fig:ABC_trn_ex}, and summarized in Table~\ref{tb:all_results}. 

%\paragraph{Straight Vine Robots.} 
Following the same pattern as before, we first look at the performance of straight vine robots to establish a point of comparison and then repeat the tests using the action space with turning. The larger obstacle size seems to create many more areas which cannot be accessed with the straight robots alone, as seen in the results of Figure~\ref{fig:ABC_str_ex}, especially the bottom row where a full obstacle was not able to be accessed. As a result, both MC and Heuristic selections perform much closer to random selections over the 11 environments, with both their average performances unable to stay out of the range of the Random performance's standard deviation (Table~\ref{tb:all_results}). In contrast, the addition of turning allows deeper exploration into the regions hidden from the action space with only straight vine robots, with the turns seeming to increase the chances of encountering a gap between obstacles that would otherwise deflect a straight vine robot. Additionally, looking at the standard deviation for these experiments, there is a much larger variance in the straight vine robots, likely due to many getting stuck in concave vertices.

\begin{figure}[t!]
\centerline{\includegraphics[width=1.\columnwidth]{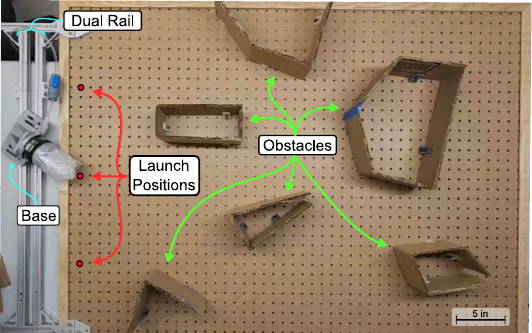}}
\caption{Setup of the physical environment with non-uniform obstacles, including a double rail for positioning and orienting the vine robot base such that deployments are accurately angled to grow through one of the three launch positions (marked in red).}
\label{fig:demo_setup}
\end{figure}

\subsection{Physical Demonstration of Mapping}
Finally, to demonstrate the power of the kinematics and their match to the real life vine robot behavior in this mapping task, we implement the proposed MC-inspired algorithm to explore an unknown environment using real vine robot deployments to generate the sensed information about the non-uniform obstacles in the environment. An overhead camera is used as the sensor, but the only information provided to the algorithm is equivalent to a tip contact sensor and length measurement, as achieved in prior work \citep{frias2023}.

\subsubsection{Physical Environment Setup}
A pegboard with 2.54~cm (1~in) spaced holes was used to create an environment with non-uniform obstacles. Cardboard walls were clipped into the holes in semi-random configurations to create polygonal obstacles. Figure~\ref{fig:demo_setup} shows the setup of the environment. For the purposes of the physical demonstration, we increased the number of launch positions to three.
Due to the length of the vine robot base, it was mounted onto a dual-rail behind the starting position, such that regardless of launch angle, the base could be positioned with its opening starting \textit{at} the marked launch position with the correct angle (Figure~\ref{fig:demo_setup}).

\begin{figure}[t!]
\centerline{\includegraphics[width=1.\columnwidth]{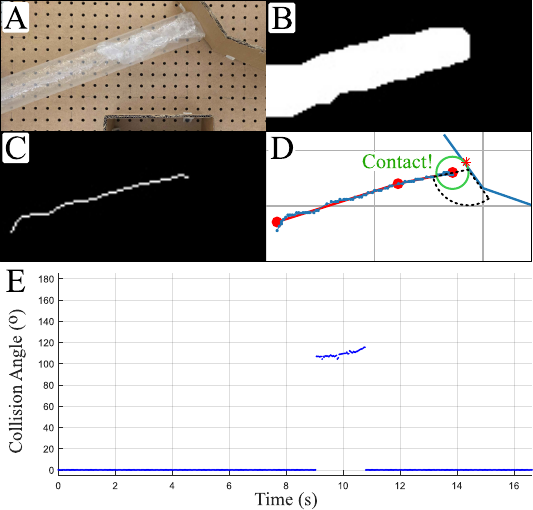}}
\caption{Example of how \textit{each} video frame is transferred into realistic sensor data streams for mapping. A) High quality image representing a frame of the video. B) The mask created comparing the video frame to a reference frame created before deployment. C) Centerline of the mask's pixels and D) piecewise linear regression of centerline to create a piecewise spline. Finally, E) the accumulation of each frame's collision angle and robot length (not pictured) creates a realistic sensor data stream over the entire deployment.}
\label{fig:irl2sim}
\vspace{-1.0em}
\end{figure}

\begin{figure}[t!]
\centerline{\includegraphics[width=0.95\columnwidth]{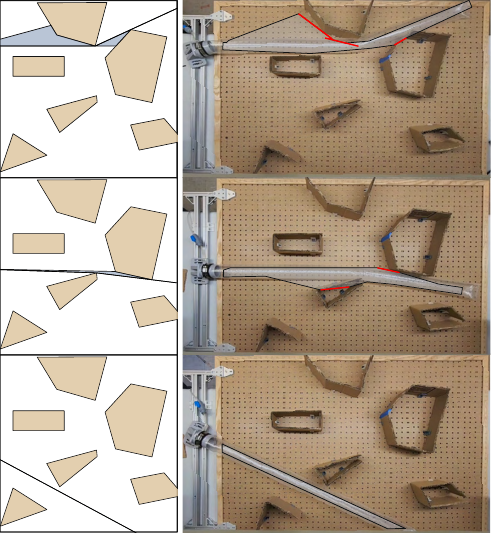}}
\caption{Three expected deployments' information in comparison to the real deployments. The rows show the fourth, sixth, and final simulated and real deployments, respectively.}
\label{fig:demo_ex}
\vspace{-1.0em}
\end{figure}

\begin{figure*}[t!]
\centerline{\includegraphics[width=2\columnwidth]{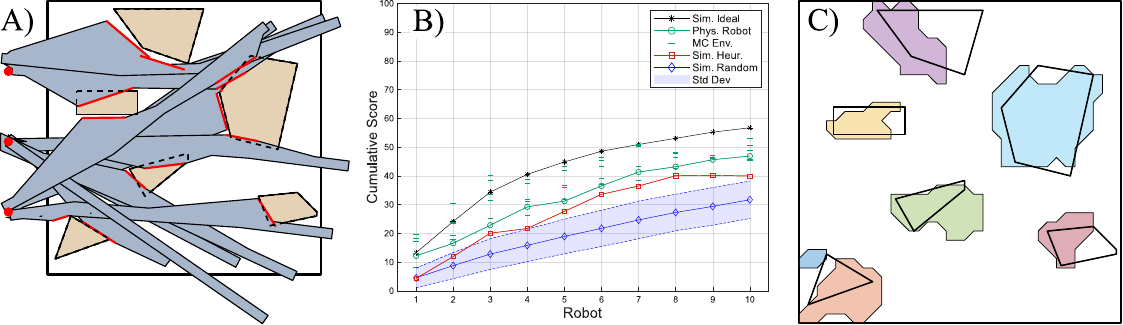}}
\caption{Results of the physical experiment. A) Wall collision (red) and swept area (blue-gray) data extracted from each deployment. B) Cumulative score after each robot launch (Phys. Robot) and the scores for 5 environments generated by the Monte Carlo simulation after each launch (MC Env.), compared to simulated Ideal, Heuristic, and Random selections. Decreasing spread of the Monte Carlo environment scores and grouping around the cumulative robot score indicates increasing confidence in the environment map over deployments. C) One generated MC environment from the final cumulative belief.}
\label{fig:demo_results}
\vspace{-1.0em}
\end{figure*}

\subsubsection{Measurement Generation from Overhead Video}
Although sensor design is beyond the scope of this paper, for the physical demonstration, we aimed to mimic sensing previously shown in integrated vine robots by \citet{frias2023} to show how well mapping could be achieved in a real scenario. The sensor data required to implement \Cref{eq:str_wall_guess} consists of the robot length and tip contact angle over time. To collect these measurements, a phone camera was placed 90~cm above the environment, and a video was recorded at 30 frames per second for each deployment. We note that while this allowed us to theoretically extract the full vine robot shape over time, we only use the tip contact angle and length during this demonstration.

To extract the tip contact angle and length from the deployment videos, we color segmented the video and then calculated the position and orientation of the shapes (Figure~\ref{fig:irl2sim}). The first frame was used to both manually select the wall locations, and to establish a baseline to compare against the rest of the frames, allowing for the creation of a mask of the robot location for each frame. These masks result in a white blob showing the location of the robot against a black background (Figure~\ref{fig:irl2sim}B). From there, we find the centerline of the blob, which calculates a 1~pixel wide representation of the vine robot (Figure~\ref{fig:irl2sim}C). The pixels are converted to positions and fed into a piecewise linear regression algorithm to create a geometric centerline of the vine robot across each frame of the video via key points needed to reconstruct the robot shape. With these key points the vine robot's length can be estimated at each frame by adding the length between each keypoint. An example of the the piecewise linear regression is shown in Figure~\ref{fig:irl2sim}D.

The last two key points are used to estimate when the vine robot's tip is in collision with a wall and angle of that collision. To detect collision at each frame, we calculate the distance between the vine robot tip (marked as the last key point) and each obstacle wall. If any obstacle falls within the vine robot's nominal radius, then a collision is marked (Figure~\ref{fig:irl2sim}D). Then, using the line segment created by the furthest two key shape points and the wall it contacted, the relative contact angle is calculated. To be consistent with sensor collection and analysis, the collision angle is always calculated counter-clockwise from the direction of growth of the vine robot's tip. After each frame of the deployment video is analyzed, we can generate two mimicked sensor data sets: vine robot length and tip collision angle (Figure~\ref{fig:irl2sim}E). The data sets were checked manually to remove any points with false contact detections, and a low pass filter was applied.

\subsubsection{Experiments and Results}
With these two mimicked sensor data sets from the real deployment scenarios, the equations from Section~\ref{sect:kin_sens} were applied to calculate the location of contacted walls and swept areas, with respect to the known launch position and angle. We then used this gathered information within the MC-inspired algorithm to estimate the best vine robot to deploy next to maximize knowledge of the environment. This process of deployment, video analysis, and Monte Carlo simulation were repeated a total of 10 times. While this is fewer than the previous experiments, the width of the robot make this effectively a much denser environment, drastically increasing the information gained from each deployment. Figure~\ref{fig:demo_ex} shows various snapshots of these deployments, both in simulation and in reality.

After the 10 deployments and their respective analysis, we review the overall performance of the algorithm when applied to a physical environment. The superposition of each deployment's analysis using Section~\ref{sect:equations}'s equations is shown in Figure~\ref{fig:demo_results}A, with estimated wall collisions in red and estimated swept areas in the blue-grey. The cumulative score after each launch, shown in Figure~\ref{fig:demo_results}B, is calculated in the same manner as was done for the simulated experiments, but is labeled Phys. Robot to distinguish that this arises from the physical launch of the robot, not simulation. The comparable Ideal, Random, and Heuristic scores based on simulation of this environment are included for comparison. Additionally, the cumulative score is compared to the scores of half of the random environments generated as part of the Monte Carlo simulation after each launch, scored using the same rubric applied to the cumulative belief matrix. These environment scores have quite a large spread early on, but in the final three launches the scores converge with less variance around the cumulative score, indicating higher certainty about the environment. Figure~\ref{fig:demo_results}C shows one of the environments generated after the final launch, confirming this close match between generated and true environments.

Finally, after the last deployment and its analysis, the final belief is used to estimate the structure of the environment (Figure~\ref{fig:demo_results}C). The results show that the obstacles' general areas are located well, even with the noise expected in a realistic deployment.

\section{Discussion}
This paper developed and tested three novel advancements for soft growing robots that build upon each other to demonstrate the feasibility of using vine robots as tools for exploring unknown environments. These contributions advance the state of the art for leveraging compliance to sense features of environments in new ways.

\subsection{Kinematic Modeling}
The results of testing~\Cref{Eq:lean_static_alt,Eq:static_squish,Eq:squish_lean}, shown in~\Cref{fig:morph_exp}, demonstrate that they can be used to reliably predict the buckling form and direction of a growing continuum structure. 
The primary source of error in these equations, as well as the ones in \citet{Haggerty2019}, occurs in the transition region between slide directions---the Friction-Locked regime identified in this work. This region is heavily influenced by friction, with higher friction leading to a larger transition region, and thus a larger region of increased uncertainty in predicting the final slide direction. This understanding leads to a general suggestion that when using passive deformation to control a growing continuum structure navigating a known environment, to avoid collisions in this region in order to have a high degree of confidence in localization and future behavior. If the environment is unknown, though, methods to reduce friction and/or to actively change the collision angle can work to avoid this uncertainty. 

\subsection{Geometric Simulation}
We see that this is similarly true in the construction and testing of the simulator. 
The largest prediction errors occurred exclusively due to Friction-Locked behavior in both straight and turning robot collisions. Outside of these collision events, the simulator is able to accurately estimate the trajectory, reinforcing our suggestion to avoid these regions of uncertainty where able. Errors were also largely bounded by the robustness of the observation that the vine robot path follows the environment's visibility graph. 

\subsection{Passive Deformations for Mapping}
Across both the uniform and non-uniform environments, the introduction of turning into the action space increased the final scores of each of the four approaches (Ideal, Monte Carlo, Heuristic, and Random) to mapping. However, an important observation is that the \textit{magnitude} of the improvements is vastly different for each method. Looking to Table~\ref{tb:all_results}, which summarizes the results of every environment and action space tested, we see that the introduction of turning to the action space is more pronounced in the non-uniform environments than the uniform environments, indicating that turning may have a stronger effect on mapping peculiarly-shaped concave obstacles and denser maps, perhaps due to this creating larger regions which are inaccessible with only straight robots. We also note that the Monte Carlo approach increased performance in both uniform and non-uniform environments by more than the Random approach did when turning was added, 6.44 and 9.70 compared to 4.67 and 7.40. Despite a sixfold increase in available actions, our algorithm is able to consistently perform better than purely random selections, including taking better advantage of the new turning actions and increased action space. Though our Heuristic is able to perform as well as the Monte Carlo approach on some environments, the performance was the most variable of all methods and only slightly better than Random on average. Overall, the MC-algorithm performance is able to exceed the expected performance of the Random policy within 5-10 deployments, directly demonstrating the algorithm building knowledge as it selects deployments. In a number of environments, the addition of turning caused that point where the MC-algorithm does better than Random to occur 3 deployments earlier, reinforcing the idea that introducing turning into the action space significantly improves mapping performance.

Finally, we note that in this mapping algorithm we are making use of standard methods which are not necessarily optimized for our robot or sensing approach, such as the grid discretization. As a result, many of the MC-generated environments have rough features  or outcroppings on sensed walls, which visually may appear to be a significant error. However, these features actually highlight the strengths of the approach, as the kinematics are still able to stably predict the expected behavior under contact even with relatively low resolution estimates of the environments, leading to consistently good action selection. This performance is further exemplified in the physical demonstration, which showed significant mapping in a real environment. Even with noise in the pseudo-measurements of the collision angles, the algorithm is still able to make educated decisions and aim toward regions of uncertainty to increase its understanding of the environment, showing robustness in its performance, \textit{despite} fewer deployments. With more tailored algorithmic approaches, mapping with vine robot deformations could quickly and reliably sense unstructured and cluttered environments. 

\subsection{Potential for Integrated Sensing}

The primary limitation of our current work is the lack of integrated sensing in our physical vine robot mapping demonstration. However, the kinematic equations allow translation of proprioceptive to exteroceptive and vice versa, meaning that a wide array of sensing strategies could be integrated with our algorithms. Two highly likely avenues towards integrated sensing are suggested by existing vine robot literature, either making proprioceptive or exteroceptive measurements. As discussed previously, and imitated by our physical demonstration, \citet{frias2023} measured tip contact angle and robot length, though with limited resolution. Future work would need to increase the resolution of this sensing method to get accurate mapping. Alternatively, as shown by \citet{gruebele2021distributed}, we could integrate distributed shape sensing of the vine robot using gyroscopes. Previous work has shown that this results in fairly accurate sensing in constrained, pipe-like environments, though with some sensor drift.

\section{Conclusions and Future Work}
Actively leveraging compliance found in soft robots, instead of passively benefiting from it, is a challenge that requires improvements in a range of topics including modeling, control, simulation, and planning, to name a few. In this paper, we took preliminary steps towards realizing more effective soft robot systems by tackling modeling and simulation challenges in enabling soft growing robots to be used as environment sensing tools.

The previously established and newly derived kinematic equations for vine robots highlight the power of compliant responses with relatively simple modeling. By extending the kinematics to include vine robots with pre-planned turns, without losing the predictive power of the previous model where discrete behaviors are decided by relatively simple geometric features, we show that we are able to move towards designing vine robots which can pre-plan or reactively-plan passive responses to achieve desired tasks in its deployment. These kinematics then lead into the observations that vine robots follow some subset of edges in an environment's visibility graph, which allowed us to consistently simulate the trajectories taken by vine robots into any given 2D environment. The direct outcome of these modeling and simulation results is shown in the performance of our Monte Carlo-inspired algorithm, which is able to sequentially map environments, consistently select better than random actions, and leverage vine robots with turns to reach previously unreachable areas.

Our next steps in the realm of collision kinematics of growing continuum structures will see the inclusion of continuous-type turns. We speculate these new equations will arise from a series of moment equations at each micro buckle on the structure. Additionally, by extending the analysis to include the third, vertical, dimension of potential movement, we can plan for turns to also be used for going \emph{over} obstacles, not just around.

In the physical demonstration section, we mimicked sensor readings using an overhead camera. In real deployments in the field, we would not have this kind of setup, so in future work, we aim to develop and incorporate an integrated sensor that will provide accurate measurements of the vine robot to allow our mapping algorithm to work in realistic scenarios. This move will also require us to consider how sensor noise affects the robustness of our mapping approach, so future work will investigate belief update methods which accommodate sensor noise and uncertainty. Another limitation is the current approach's effectiveness may diminish in significantly larger environments where single turns provide insufficient reach. Future work should investigate multi-turn strategies and adaptive turning for larger-scale exploration tasks.

\section*{\textbf{Acknowledgments}}
The authors would like to acknowledge Lucas Chen and Yitian Gao for their assistance in the code created for extracting the mimicked data set.

\section*{\textbf{Author Contributions}}
F.F. and L.H.B. conceived of the presented ideas. Z.K. contributed to software and conceptual methodologies. F.F. designed and performed experiments. F.F. and S.D analyzed data.  F.F., Z.K., and L.H.B. wrote the original draft and edited the manuscript. All authors approved of the final paper.

\section*{\textbf{Statements and Declarations}}

\subsection*{Ethical Considerations}
This article does not contain any studies with human or animal participants.

\subsection*{Consent to Participate}
Not applicable.

\subsection*{Consent for Publication}
Not applicable.

\subsection*{Declaration of Conflicting Interests}
The author(s) declared no potential conflicts of interest with respect to the research, authorship, and/or publication of this article.

\subsection*{Funding Statement}
The authors disclosed receipt of the following financial support for the research, authorship, and/or publication of this article: This work was supported by National Science Foundation Graduate Research Fellowship Program [DGE-2444108].

\bibliographystyle{SageH}
\bibliography{main}

\end{document}